\documentclass[runningheads]{llncs}
\usepackage{CJKutf8}
 
\usepackage{eccv}

\usepackage{eccvabbrv}

\usepackage{graphicx}
\usepackage{booktabs}
\usepackage{pifont}
\usepackage{tabularx}
\usepackage{array}
\usepackage{tcolorbox}
\tcbuselibrary{breakable}
\usepackage{listings}
\newcommand{\cmark}{\ding{51}}

\usepackage[table]{xcolor}

\definecolor{BestColor}{HTML}{E2F0D9}    
\definecolor{SecondColor}{HTML}{FFF2CC}
\newcommand{\best}[1]{\cellcolor{BestColor}\textbf{#1}}
\newcommand{\second}[1]{\cellcolor{SecondColor}#1}

\newcommand{\bestbox}{\colorbox{BestColor}{\phantom{10}}}
\newcommand{\secondbox}{\colorbox{SecondColor}{\phantom{10}}}

\usepackage[accsupp]{axessibility}  

\usepackage{hyperref}

\usepackage{orcidlink}
\usepackage{marvosym}

\begin{document}
\begin{CJK*}{UTF8}{gbsn}

\title{
HanMoVLM: Large Vision-Language Models for Professional Artistic Painting Evaluation
} 

\titlerunning{HanMoVLM}


\authorrunning{Yang et al.}

\def\spaces{}
\author{Hongji Yang\inst{1} \and Yucheng Zhou\inst{1} \and Wencheng Han\inst{2} \and Songlian Li\inst{3} \and  \\ Xiaotong Zhao\inst{3} \and Jianbing Shen\inst{1}\textsuperscript{\Letter}
}

\institute{SKL-IOTSC, CIS, University of Macau \and CSE, Shandong University \and Online-Video BU, Tencent
}

\maketitle

\def\thefootnote{{\text{\Letter}}}\footnotetext{
Corresponding author.
}

\begin{abstract}
  While Large Vision-Language Models (VLMs) demonstrate impressive general visual capabilities, they remain artistically blind and unable to offer professional evaluation of artworks within specific artistic domains like human experts. 
  To bridge this gap, we transform VLMs into experts capable of professional-grade painting evaluation in the Chinese Artistic Domain, which is more abstract and demands extensive artistic training for evaluation.
  We introduce \textbf{HanMo-Bench}, a new dataset that features authentic auction-grade masterpieces and AI-generated works, grounded in real-world market valuations. To realize the rigorous judgment, we propose the \textbf{HanMoVLM} and construct a Chain-of-Thought (CoT) validated by experts. This CoT guides the model to perform expert-level reasoning: from content identification and Region of Interest (RoI) localization to professional evaluation, guided by both theme-specific evaluation and typical three-tier evaluation in Chinese paintings.
  Furthermore, we design a reward function to refine the reasoning process of the HanMoVLM to improve the accuracy. We demonstrate that HanMoVLM can serve as a critical backbone for Test-time Scaling in image generation. By acting as a high-quality verifier, HanMoVLM enables generative models to select the most artistically superior outputs from multiple candidates. Experimental results and human studies confirm that the proposed HanMoVLM effectively bridges the gap, achieving a high consistency with professional experts and significantly improving the quality of Chinese Painting generation.
\end{abstract}

\begin{figure*}[t]
    \centering
    \includegraphics[width=1.0\linewidth]{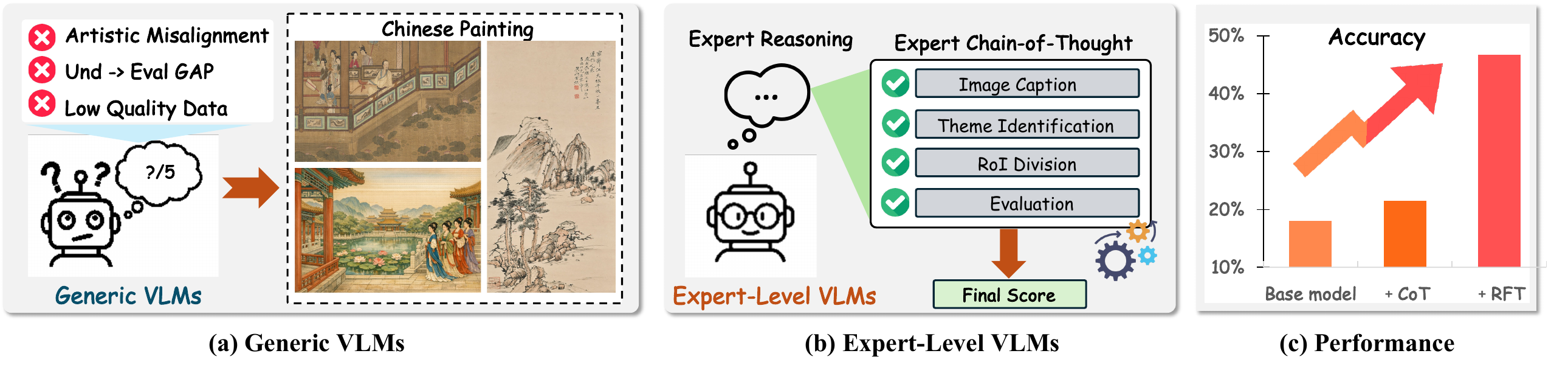}
    \vspace{-6mm}
    \caption{The motivation of our study. \textbf{(a)} Generic VLMs suffer from artistic misalignment, evaluation gaps, and low-quality data when handling domain-specific artistic content such as Chinese paintings; \textbf{(b)} The expert-level VLMs introduce the structured expert chain-of-thought, to produce reliable final scores; \textbf{(c)} Experimental results show that our CoT and RL significantly improve performance. }
    \vspace{-6mm}
    \label{fig:motivation}
\end{figure*}

\section{Introduction}
\label{sec:intro}

Large Vision-Language Models, in recent years, have demonstrated extraordinary capabilities in general visual understanding and logical reasoning. Although the VLM can identify ``mountains'' or ``rivers'' across diverse artistic styles in paintings, including Chinese paintings, it still lacks the domain-specific artistic reasoning required to evaluate non-Western visual art~\cite{liu2025culturevlm, zhang2025creating, yu2025structured, yu2026cross, tao2024cultural}. In this study, we take Chinese painting as the subject of investigation, due to its rich artistic context, drive itself distinct from conventional visual aesthetics\cite{zhang2025cultiverse}. 
The evaluation of Chinese paintings does not rely on realism or linear perspective, but rather a unique system of judgment standards rooted in cultural depth. 
Consequently, evaluating such artworks is not an easy thing for the layman because it requires expert-level cognition through years of training in the Chinese arts. Current generic VLMs still struggle to evaluate these paintings like the experts in this field.
As shown in Fig.~\ref{fig:motivation}, the challenges of artistic understanding remain unresolved, due to \ding{172}~\textbf{Artistic Misalignment}, where the generic model lacks understanding of Chinese painting for assessment.
\ding{173}~\textbf{Understanding-Evaluation GAP}, where the generic model can identify the elements of Chinese painting but still be unable to evaluate.
\ding{174}~\textbf{Low Quality Data}, there is a lack of high-quality annotated data that is under the review of experts.

To bridge the gap between general visual recognition and professional artistic evaluation, we present HanMoVLM~\def\thefootnote{1}\footnote{HanMo(翰墨)~refers to brush and ink, it often symbolizes scholarship and traditional Chinese artistic expression.}, which leverages the expertize of Chinese painting to assess artwork through a professional Chain-of-Thought~\cite{wei2022chain, wangself, zhouleast}.
To achieve this, we design an expert-level CoT, which begins with identifying the context and theme of the painting, and concentrating on RoIs for the details. 
Upon this, HanMoVLM can select the corresponding evaluation criteria based on the determined theme and assess the work using the three-tier evaluation method for Chinese painting. 
To ensure the correctness of CoT in reasoning and final answers, we propose a reward function that rewards each component of the chain. We employ the Group Relative Policy Optimization (GRPO)~\cite{guo2025deepseek} for Reinforcement Fine-Tuning (RFT). 
Finally, to facilitate rigorous and standardized evaluation of reasoning in VLMs, we construct a new dataset (HanMo-Bench), which contains Chinese paintings for visual artistic understanding and is reviewed by experts.
For visual artistic generation, we employ test-time scaling generation. Each text-to-image (T2I) model samples multiple outputs as candidates, and the HanMoVLM is employed as the external verifier for the Best-of-N\cite{cobbe2021training} strategy. 
Experiments on HanMo-Bench validate the effectiveness of the proposed pipeline.
Our contribution can be summarized as follows:
\begin{itemize}
    \item We introduce a new task to evaluate the capability of VLMs in assessing artistic paintings and construct a novel dataset, HanMo-Bench. This benchmark aims to assess whether current VLMs can transcend general visual recognition to achieve painting evaluation within artistic contexts.
    \item We propose HanMoVLM, a powerful VLM for Chinese painting evaluation, with a professional-level Chain-of-Thought reasoning. This CoT enables VLMs to perform structured expert reasoning by integrating global and local visual analysis, theme-specific criteria, and typical three-tier criteria. 
    \item We design a new reward function specifically aligned with the CoT. By providing granular supervision across theme identification, RoI localization, and the rationality of rating, this reward significantly enhances the model’s reliability and precision as a professional evaluator in a specific art.
    \item We empower artistic painting generation through a Test-time Scaling strategy. By utilizing HanMoVLM as an expert-level judge to select the best outputs from multiple candidates sampled by existing text-to-image models. Human studies confirm that our model is highly consistent with professional art experts.
\end{itemize}

\section{Related Work}
\label{sec:related_work}

The rapid development of vision-language models also raises concerns about artistic values and knowledge\cite{nayak2024benchmarking, romero2024cvqa, xu2025tcc}. Most VLMs and T2I models are pre-trained on internet-scale datasets that exhibit a heavy Western-centric bias, leading to the ``artistic blindness'' when encountering non-Western semantics.

\subsection{Visual Artistic Understanding}


Many efforts have been devoted to evaluating whether VLMs can accurately interpret content from different arts~\cite{yuan2023artgpt, cetinic2022understanding, hayashi2024towards, pado2025artwork}.
Representative work on artistic domain like the artistic and aesthetic dataset(SemArt~\cite{garcia2018read}, Artemis~\cite{achlioptas2021artemis}, Ava-dataset~\cite{murray2012ava}, Artimuse~\cite{cao2025artimuse}, WikiArt~\cite{saleh2015large}, and OmniArt~\cite{strezoski2018omniart}), benchmarks~\cite{alfarano2025vqart, yu2026vulca}, artistic caption~\cite{bai2021explain}, artistic interpretation~\cite{bin2024gallerygpt,pado2025artwork}, artistic evaluation~\cite{zheng2025artmentor, strafforello2025have}, proves that existing VLMs struggle to understand the specific artists (especially non-Western arts)~\cite{liu2025culturevlm, zhang2025creating, yu2025structured, yu2026cross, tao2024cultural}.
However, these studies focus on capturing artistic content or translating the artistic symbol, without any consideration of expert-level artistic evaluation within a specific artistic background.
Our study is the first to investigate this task and focus on Chinese art painting, whose evaluation requires extensive artistic training to reach the expert level~\cite{wan2024wumkg}.

\subsection{Visual Artistic Generation}

Visual artistic generation is a longstanding research topic in generative models~\cite{wang2025diffusion, ko2023large, gong2025cta, he2018chipgan}. To introduce artistic concepts, some methods\cite{zhao2025novel, yang2024artfusion, liao2023calliffusion,tang2025art} apply LoRA\cite{hu2022lora} or ControlNet\cite{zhang2023adding} to adopt a certain style into the content and thus introduce artistic concepts.
However, although generative models produce impressive visual results, the outputs do not fully align with aesthetic values shaped by specific artistic contexts. For example, the artistic value of Chinese painting does not fully derive from realism or accurate linear perspective. 
Besides, the evaluation of artistic generation still relies heavily on labor-intensive validation by human experts, which prevents timely feedback and automated evaluation during training.



\section{Method}

\subsection{Preliminary}
    
\noindent\textbf{GRPO.}
Given an input $x$, the large language model samples multiple outputs $\{o_1,o_2,\cdots, o_G\}$ from the original model $\pi_{\theta_{\text{old}}}$ and computes a group of rewards $\{r_1,r_2,\cdots, r_G\}$ through predefined reward function. Then, the advantages corresponding to the outputs are calculated as:
\begin{equation}
    A_i = \frac{r_i-\text{mean}(\{r_1,r_2,\cdots, r_G\})}{\text{std}(\{r_1,r_2,\cdots, r_G\})},
\end{equation}
Therefore, the policy model $\pi_{\theta}$ can be optimized by maximizing the objective function as follows:
\begin{equation}\small
\begin{aligned}
\mathcal{J}(\theta) = \mathbb{E}_{\begin{subarray}{l}
    \{o_i\}_{i=1}^G \sim \pi_{\theta_{\text{old}}}(O|q) \\
    q \sim P(Q) 
  \end{subarray}} \Bigg[ \frac{1}{G} \sum_{i=1}^G \bigg( \min \left( \rho_i A_i, \text{clip} \left( \rho_i, 1-\epsilon, 1+\epsilon \right) A_i \right) \Bigg].
\end{aligned}  
\end{equation}
where $\rho_i=\frac{\pi_{\theta}(o_i|q)}{\pi_{\theta_{old}}(o_i|q)}$, and $\epsilon$ is a hyperparameter. The $P(Q)$ denotes the distribution of questions in the datasets.

\begin{figure*}[t]
    \centering
    \includegraphics[width=1.0\linewidth]{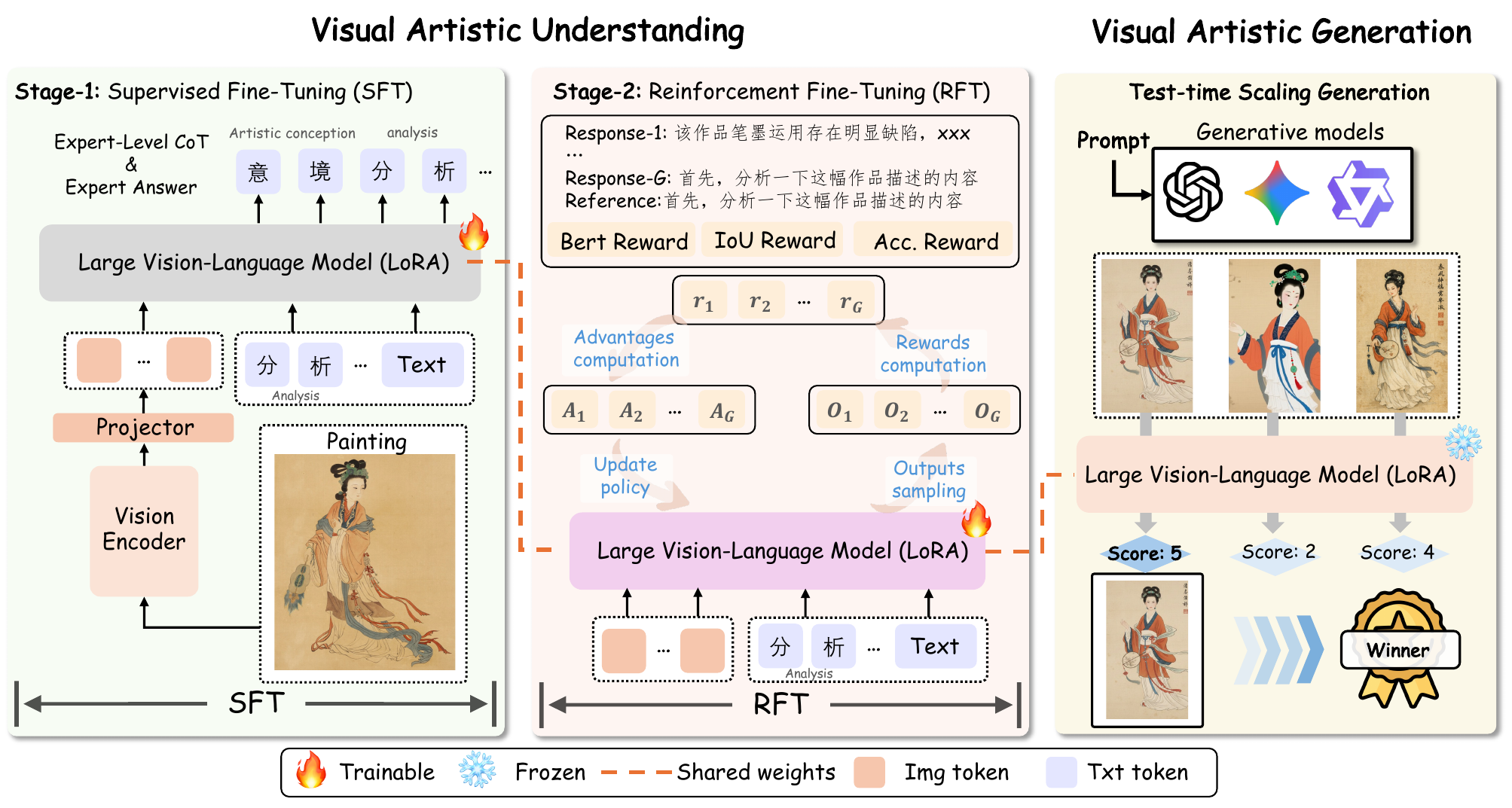}
    \vspace{-6mm}
    \caption{The overall framework of HanMoVLM. For visual artistic understanding, we finetune HanMoVLM with expert-level CoT and expert answer via SFT, and then perform RFT based on the expert reward function. For visual artistic generation, we apply the frozen HanMoVLM as the external verifier to evaluate the images sampled from existing T2I models.}
    \vspace{-4mm}
    \label{fig:overall_framework}
\end{figure*}

\begin{figure*}[t]
    \centering
    \includegraphics[width=1.0\linewidth]{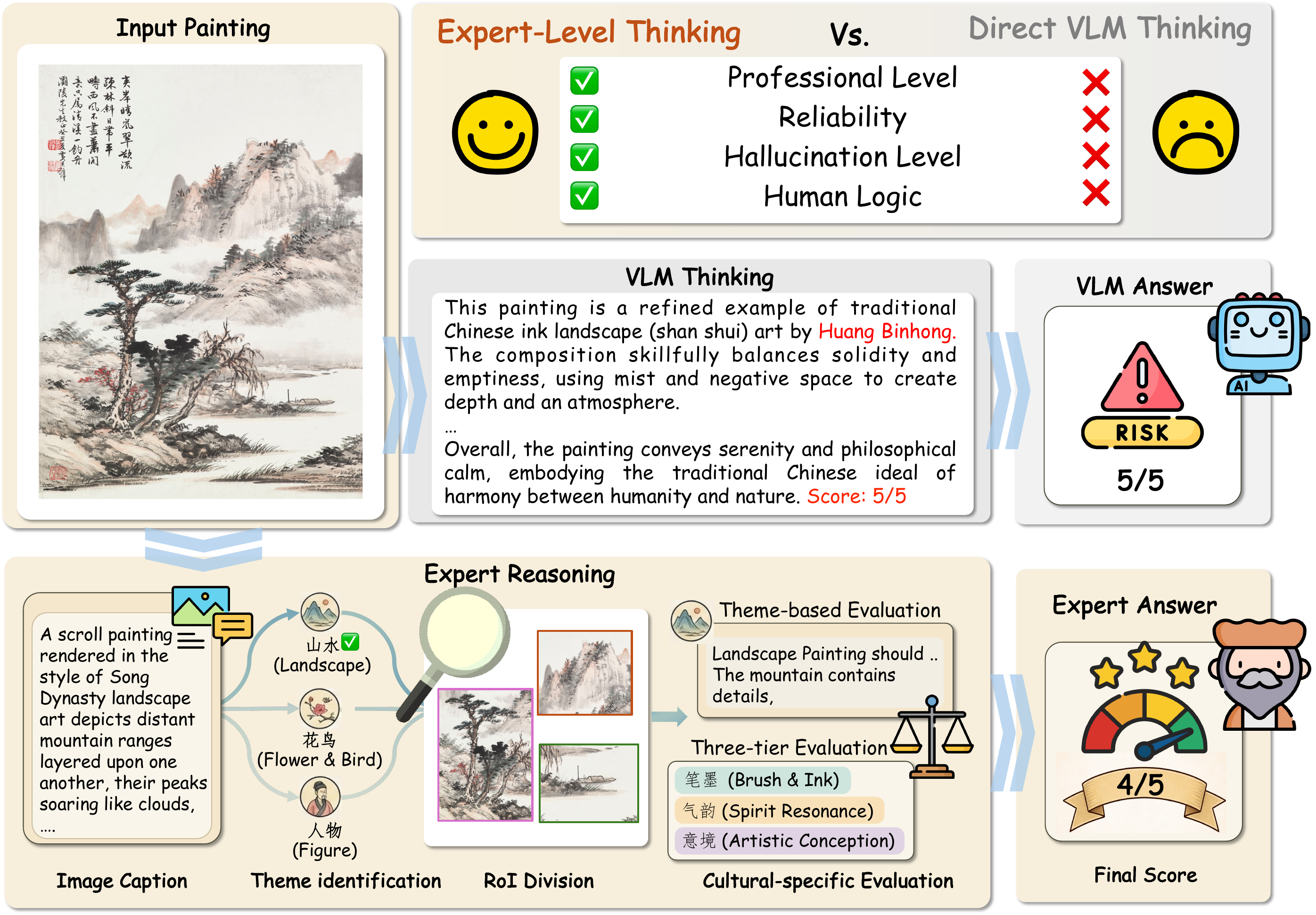}
    \vspace{-6mm}
    \caption{The expert-level Chain-of-Thought in our HanMoVLM. When the generic VLM evaluates Chinese paintings, the model tends to generate outputs based on general knowledge (non-expert) rather than following professional guidelines. As a result, its analysis does not reflect the experts' procedures, leading to a decline in terms of professional level, reliability, hallucination, and human logic.}
    \vspace{-6mm}
    \label{fig:expert_cot}
\end{figure*}

\subsection{HanMoVLM Framework}
In this section, we present the overall framework of HanMoVLM, which can serve as an evaluator in artistic painting evaluation and the external verifier in artistic painting generation. As illustrated in Fig.~\ref{fig:overall_framework}, the proposed method consists of two training stages, SFT and RFT, to integrate the expert reasoning and responses of paintings in HanMoVLM. 
The model takes painting as visual input and produces structured textual analysis, enabling both semantic understanding and aesthetic evaluation.
Therefore, it mitigates artistic misalignment and bridges the gap between understanding and evaluation. 
In addition, HanMoVLM can be also employed for artistic generation, serving as an evaluator to assess the quality of generated images.

\subsection{Artistic Expert-Level Chain-of-Thought}
\label{sec:cot}
Our core idea is to design a CoT that mimics the perspective of Chinese art experts. To realize an expert-level CoT, we first establish a structured evaluation procedure based on the guidance of experts, which follows a progression from global understanding to detailed assessment as shown in Fig.~\ref{fig:expert_cot}.
Specifically, HanMoVLM first identifies the overall content of the given painting and determines its precise theme within the system of Chinese painting in Landscape(山水), Flower\&Bird(花鸟), and Figure(人物). Then, the HanMoVLM conducts a detailed analysis of the main visual elements or the local regions of interest, focusing on critical details and primary visual focal points. After completing global and local assessments, the HanMoVLM selects the appropriate evaluation criteria according to the theme of the painting, as the assessment standards vary significantly across different themes. Finally, the painting is evaluated based on the typical three-tier criteria of Chinese painting, which are Brush\&Ink(笔墨), Spirit Resonance(气韵), and Artistic Conception(意境). 
After integrating all these professional analyses, the HanMoVLM produces the final score.

\subsection{Expert Reward Function}
After integrating expert reasoning into the CoT, a primary challenge is to prevent the VLM from generating hallucinated reasoning and incorrect responses based on the CoT. To address this issue, we design a set of reward functions that assign to different components of the reasoning chain (as described in Sec~\ref{sec:cot}), as well as the final output score, during the RFT process.

\noindent\textbf{Accuracy Reward.} To ensure the final score is consistent with the ground-truth, we calculate the difference between them, as follows:
\begin{equation}
    R_\text{acc} = 1 - \frac{|pred\_score-gt\_score|}{5},
\end{equation}

\noindent\textbf{BERT Reward.} 
To measure the semantic similarity of the specific part between sampled responses $\hat{x}$ and reference responses $x$, we encode the parsed $K$ parts using a pre-trained BERT model\cite{zhangbertscore}. This process can be formatted as follows:
\begin{equation}
    R_\text{BERT}=\frac{1}{K}\sum^K_{j=1}\text{BERTScore}(pred_j, gt_j)
\end{equation}

\noindent\textbf{IoU Reward.} 
In addition, we compute the mIoU between the predicted RoIs and the ground-truth bounding boxes in the reference responses. For each predicted box, we match it with the gt boxes that achieve maximum IoU as its corresponding target. Moreover, since this task is an open-vocabulary detection, we also compute the BERTScore for the description in each pair of boxes.

\begin{equation}
    j^*(i) = \arg\max_j IoU(pred\_box_i, gt\_box_j),
\end{equation}
\vspace{-5mm}
\begin{equation}
    R_\text{mIoU} = \frac{1}{N}\sum^N_{i=1}\Big[IoU(pred\_box_i, gt\_box_{j^*(i)}) + \text{BERTScore}(pred_i, gt_{j^*(i)})\Big],
\end{equation}
Therefore, the overall reward function, $R_{final}$, is a weighted sum of all the rewards:
\begin{equation}
    R_{final} = w_\text{acc}\cdot R_\text{acc} + w_\text{BERT}\cdot R_\text{BERT} +w_\text{mIoU}\cdot R_\text{mIoU}.
\end{equation}
where the $w_\text{acc}$, $w_\text{BERT}$ and $w_\text{mIoU}$ are the coefficients of the weight in different rewards.

\subsection{Test-time Scaling for Visual Artistic Generation}
Once the VLM is trained to provide expert-level reasoning and answers, it can act as a powerful external verifier for T2I models.
To improve the quality of the result of artistic generation, we introduce a test-time scaling strategy, which requires no additional training or fine-tuning of the generative model and achieves consistent performance gains by increasing computation only at inference time.

Given a textual input $x$, the T2I model $G_\theta$ generates $N$ sampled paintings $y$ under the fixed parameter $\theta$, which can be formatted as:
\begin{equation}
    \{y_1, y_2, \cdots, y_N\} = G_\theta(x),
\end{equation}
The VLM then takes the generated image $y$ and outputs a scalar compatibility score, which can be formatted as follows:
\begin{equation}
    s_i = \text{VLM}(y_i),
\end{equation}
where $s_i$ is the expert response rating from HanMoVLM.
The final generated image $y^*$ is selected by maximizing $s_i$ as follows:
\begin{equation}
    y^{*} = \arg\max_{y_i\in{\{y_1,y_2,\cdots, y_N}\}}\text{VLM}(y_i).
\end{equation}
This select process effectively performs distributional search guided by a multi-modal semantic objective, significantly enhancing the quality and artistic correctness of the T2I outputs.

\section{HanMo-Bench}
\label{sec:hanmobench}
\begin{figure*}[t]
    \centering
    \includegraphics[width=1.0\linewidth]{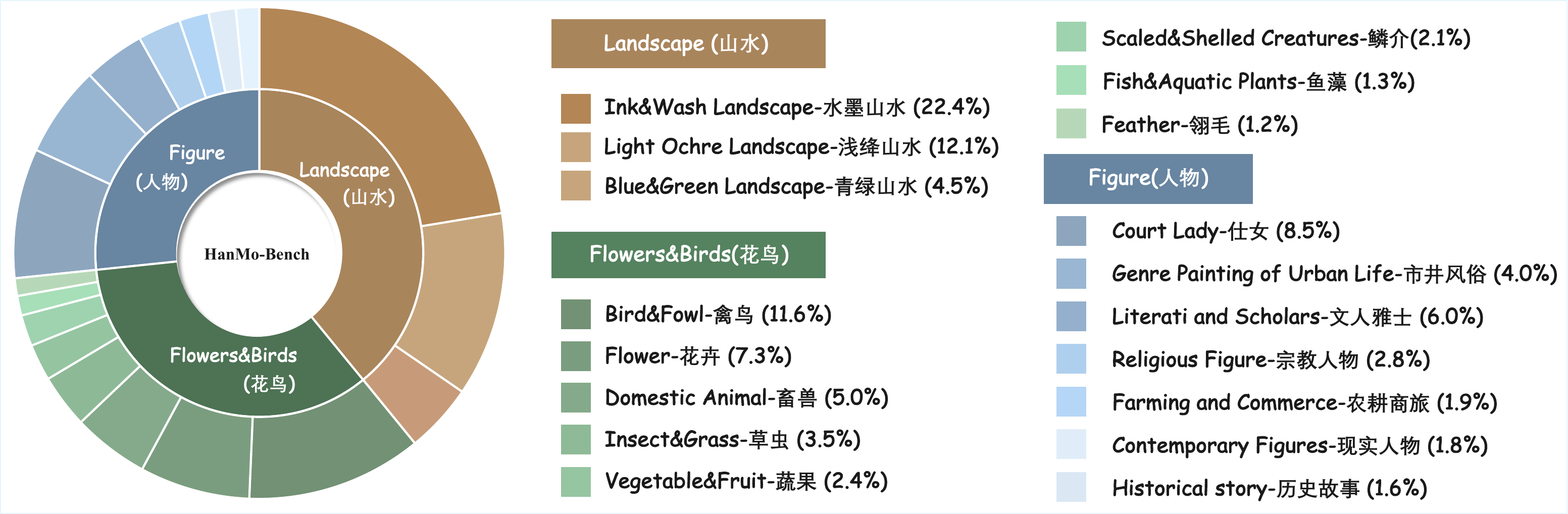}
    \caption{Composition of HanMo-Bench.}
    \vspace{-4mm}
    \label{fig:dataset_composition}
\end{figure*}

\begin{figure*}[t]
    \centering
    \includegraphics[width=1.0\linewidth]{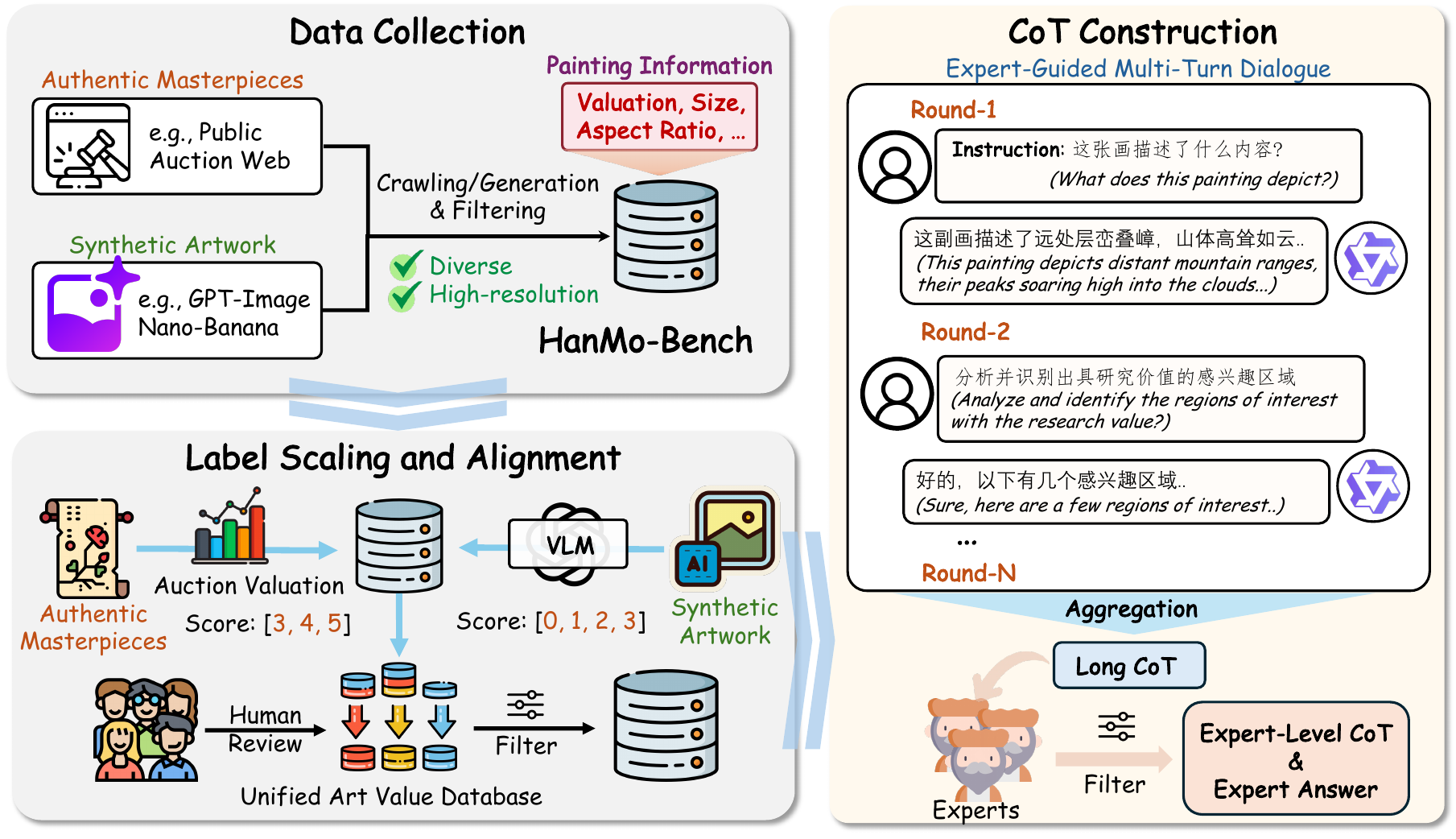}
    \vspace{-6mm}
    \caption{The construction pipeline of HanMo-Bench.}
    \vspace{-4mm}
    \label{tab:hanmobench}
\end{figure*}

To facilitate a professional evaluation of Chinese Painting, we create a novel dataset, named HanMo-bench. This dataset contains 13k Chinese paintings across different themes (Landscape, Flowers\&Birds, and Figure), as shown in Fig.~\ref{fig:dataset_composition}.
The trainset and testset contain 13,162 and 600 paintings, respectively.
The construction pipeline consists of the three stages under human validation: Data Collection, Label Scaling and Alignment, CoT construction, as shown in Fig.~\ref{tab:hanmobench}. The HanMo-Bench will be available.

\subsection{Data Collection}
To ensure the representation of both historical artistic value and the generative capabilities of modern models, we collect data from two primary sources.

\noindent\textbf{Authentic Masterpieces}
We collect high-resolution images of authentic Chinese paintings from publicly available auction house records, along with their professional valuations. These data are the representation of high artistic value Chinese paintings.

\noindent\textbf{Synthetic Artwork}.
We first employ the LLMs (GPT-5.1\cite{chatgpt_communication} and Gemini-3-Flash\cite{gemini3_communication}) to generate a diverse set of prompts in Chinese painting for T2I models. Specifically, we ask LLMs to emphasize traditional Chinese artistic styles, detailed visual elements, painting techniques, and thematic content, thereby improving the quality of generated Chinese paintings. Then, we generate images with various aspect ratios using GPT-Image-1.5\cite{gpt_image} and Nano-Banana\cite{comanici2025gemini}.

\subsection{Label Scaling and Alignment}
To bridge the gap between valuation and generative quality, we scale the value of these paintings, ranging from 0 to 5. For \textbf{Authentic Masterpieces}, we normalize their auction valuations into three tiers, assigning scores of 3, 4, and 5 to the top $60\sim100\%$, top $10\sim60\%$, and top 10\%, respectively, representing different levels of artistic and historical significance. This is because the auctioneer's valuation process is actually a human assessment and reflects the value of the paintings.
For \textbf{Synthetic Artwork}, we apply a VLM (Gemini-3-Pro\cite{gemini3_communication}) to categorize the outputs into four quality levels (0-3). This process has been reviewed by human experts. Note that we have discarded some samples to ensure a balanced sample ratio. The overlapping at score 3 is designed to simulate the current reality, where high-quality AI-generated images have attained a level approaching that of some early masterpieces, thereby increasing the difficulty for models to distinguish them.
In this way, we unify the authentic masterpieces and synthetic artwork within a unified art value system, facilitating later analysis and evaluation for VLM.

\vspace{-1mm}
\subsection{Chain-of-Thought Construction}
\vspace{-1mm}
To construct professional reasoning, we introduce a structured CoT generation using Qwen-Max\cite{bai2025qwen3} via multi-turn dialogues. The VLM is pre-conditioned on the ground-truth score and authenticity to ensure the generated reasoning is consistent with the final evaluation. The final CoT is concatenated with image caption, ROI, theme identification, theme-specific evaluation, typical three-tier evaluation on Chinese painting, and the final rating.

\vspace{-1mm}
\subsection{Human-in-the-loop Expert Validation}
\vspace{-1mm}
To ensure the quality and accuracy of the generated CoT, we implement a rigorous human-in-the-loop validation. Three experts specializing in Chinese art are tasked with evaluating the reasoning process and filtering out data that does not meet the requirements.
The label scaling and alignment are also reviewed by human to ensure the artistic value.
The HanMo-Bench will be available.

\section{Experiment}

\vspace{-1mm}
\subsection{Experimental Setup}
\vspace{-1mm}
\noindent\textbf{Training.}
Experiments are conducted on the open source base model Qwen3-VL-Instruct\cite{bai2025qwen3} in 4B and 8B versions with 8 NVIDIA H20 96GB GPUs. We employ LoRA training with a rank of 64 and an alpha of 64, respectively. For SFT, we train the model for 3 epochs with a batch size of 16 without weight decay. The initial learning rate is set to 1e-5. 
For RFT, we train our model with an initial learning rate of 1e-6 and a batch size of 32 for 2,000 steps. The number of samples for each prompt is set to 8. Please refer to supplementary for details and prompt formats.

\noindent\textbf{Artistic generation.}
For generations, we adopt the test-time scaling strategy for training-free T2I generation. We apply the Best-of-N mechanism for each T2I model to sample multiple outputs and then employ our HanMoVLM to select the highest. The number of $N$ samples is set to 8. We choose the leading T2I models (i.e., Nano-Banana\cite{comanici2025gemini}, GPT-Image-1.5\cite{gpt_image}, Seedream-4.0\cite{seedream2025seedream}, Qwen-Image\cite{wu2025qwen}) for sampling.

\begin{table*}[t]
    \caption{The comparison of our HanMoVLM with other models. \bestbox~and \secondbox~indicate the best and second best results, respectively.}
    \vspace{-4mm}
    \setlength{\tabcolsep}{8pt}
    \centering
    \resizebox{\textwidth}{!}{
    \setlength{\tabcolsep}{8pt}
    \begin{tabular}{@{}lc wc{2cm} wc{2cm} wc{2cm} wc{2cm}} 
    \toprule
    \textbf{Model} & \textbf{~Size~} & \textbf{$\text{MAE}\downarrow$}  & \textbf{$\text{RMSE}\downarrow$} & \textbf{$\text{Accuracy(\%)}\uparrow$} & \textbf{$\text{BERTScore}\uparrow$} \\ \midrule
    \rowcolor{gray!25}\multicolumn{6}{c}{Proprietary Models} \\
    \midrule
     Gemini-3-Flash\cite{gemini3_communication} & - & 0.9883 & 1.3472 & 31.17 & -- \\
    ~~~~+\textit{Thinking} & - & 0.9633 & 1.4355 & 34.83 & 0.3052 \\
    ~~~~+\textit{Expert CoT} & - & \second{0.9466} & 1.3351 & \best{37.00} & 0.3035 \\
    Gemini-3-Pro\cite{gemini3_communication} & - & 0.9866 & 1.3589 & 33.83 &-- \\
     \quad + \textit{Thinking} & - & 0.9783 & \second{1.3311} & 35.17 & 0.2855 \\
     \quad + \textit{Expert CoT} & - & \best{0.9416} & \best{1.3297} & \second{36.83} & \best{0.3101} \\
    
     GPT-4\cite{achiam2023gpt} & - & 1.8023 & 2.2387 & 17.06 & -- \\
     ~~~~+\textit{Expert CoT} & - & 1.7491 & 2.1683 & 17.41 & 0.3049 \\
     GPT-5.1\cite{chatgpt_communication} & - & 1.7742 & 2.2123 & 17.73 & -- \\
     ~~~~+\textit{Thinking} &  - & 1.8204 & 2.5200 & 17.95 & 0.1648 \\
     ~~~~+\textit{Expert CoT} &  - & 1.7671 & 2.1471 & 18.41 & \second{0.3073} \\
    
    \midrule
    \rowcolor{gray!25}\multicolumn{6}{c}{Open-Source Models} \\
    \midrule
    
    InternVL-3.5\cite{wang2025internvl3} & ~4B & 1.9591 & 2.4382 & 13.83 & 0.3441 \\
    Qwen3-VL-Instruct\cite{bai2025qwen3} & ~4B & 1.9734 & 2.4055 & ~7.90 & -- \\
    \quad + \textit{Thinking} & ~4B & 1.9493 & 2.4465 & 17.89 & 0.2836 \\
    ~~~~+\textit{Expert CoT} & ~4B & 1.9341 & 2.4495 & 19.44 & 0.2919  \\
    \textbf{HanMoVLM (SFT)} & ~4B & \second{0.6850} & \second{0.9210} & \second{39.17} & \second{0.5506} \\
    \textbf{HanMoVLM (RFT)} & ~4B & \best{0.6632} & \best{0.9002} & \best{42.50} & \best{0.5568} \\ \midrule

    STEP3-VL\cite{huang2026step3} & 10B & 2.2950 & 2.7823 & 15.67 & 0.2335 \\
    GLM4.6V-Flash\cite{hong2025glm} & ~9B & 2.0495 & 2.5466 & 15.50 & 0.3517 \\
    InternVL-3.5\cite{wang2025internvl3} & ~8B & 1.8866 & 2.3579 & 17.83 & 0.3395 \\
    Qwen3-VL-Instruct\cite{bai2025qwen3} & ~8B & 1.5582 & 1.9534 & 18.53 & -- \\
    \quad + \textit{Thinking} & ~8B & 1.5050 & 1.9370 & 19.00 & 0.3402 \\
    ~~~~+\textit{Expert CoT} & ~8B & 1.5124 & 1.9476 & 21.50 & 0.3757 \\
    \textbf{HanMoVLM (SFT)} & ~8B & \second{0.6450} & \second{0.9027} & \second{43.67} & \second{0.5531} \\
    \textbf{HanMoVLM (RFT)} & ~8B & \best{0.6310} & \best{0.8912} & \best{46.67} & \best{0.5598} \\

    \bottomrule
    \end{tabular}}
    \vspace{-6mm}
    \label{tab:main_exp}
\end{table*}

\vspace{-2mm}
\subsection{Evaluation Metrics}
\vspace{-2mm}
To further investigate the accuracy and analysis capability of the HanMoVLM, we apply multiple metrics in this task.
For the final score prediction, we use the accuracy to assess whether the VLM can correctly assign each input painting to its grade. Besides, since this task is actually a regression problem rather than a classification task, we employ Mean Absolute Error (MAE) and Root Mean Square Error (RMSE) for evaluation. 
When computing accuracy for the cases of (0 vs. 5) and (4 vs. 5), both predictions are treated as incorrect. However, the prediction of 4 is clearly closer to the ground-truth score of 5, so the MAE and RMSE are applied to reflect the gap between the true values and the estimated scores.
For models using expert-level CoT, we also report the BERTScore between the response and the reference answer to verify whether there are similar reasoning processes.
Finally, we evaluate the mIoU for RoI localization and the accuracy of theme classification, as well as the BERTScore in each RoI.

\vspace{-1mm}
\subsection{Main Results}
\vspace{-1mm}
\noindent\textbf{Visual Artistic Understanding.} To validate the effectiveness of expert-level CoT and reward functions in HanMoVLM, we compare our methods with other existing models, including Gemini-Family\cite{gemini3_communication}, GPT-Family\cite{chatgpt_communication}, InternVL\cite{wang2025internvl3}, STEP3-VL\cite{huang2026step3}, and GLM4.6V\cite{hong2025glm}. As shown in Tab.~\ref{tab:main_exp}, the expert-level CoT leads to different performance gains across various models.
For example, it exhibits a relative improvement of 18.70\% on Gemini-3-Flash and 146.07\% on Qwen-3-VL-4B-Instruct. This suggests that the expert-level CoT effectively guides the VLMs to understand and analyze Chinese paintings from an artistic perspective and eventually outputs the corresponding score.
Furthermore, while both involve reasoning, different thinking paradigms still yield performance gains. For example, Gemini-3-Flash using expert CoT improves accuracy by 3\% compared to the one perform its own thinking.
In addition, the MSE and RMSE values for GPT-Family and several models are relatively high, which indicates a significant discrepancy between the predicted ratings and the ground-truth ratings.

In addition, SFT and RFT with the designed reward function significantly improve the performance of the base model. For example, HanMoVLM-8B has a 3\% improvement in terms of accuracy. Moreover, the BERTScore improved by 49\%, indicating that the model's reasoning on the painting is increasingly aligning with expert interpretations.
We also report the improvement of fine-tuning on RoI division and description, as shown in Tab.~\ref{tab:iou_bert_theme}. The fine-tuned model demonstrates greater alignment with experts in segmenting and describing RoIs.

\begin{figure*}[t]
    \centering
    \includegraphics[width=1.0\linewidth]{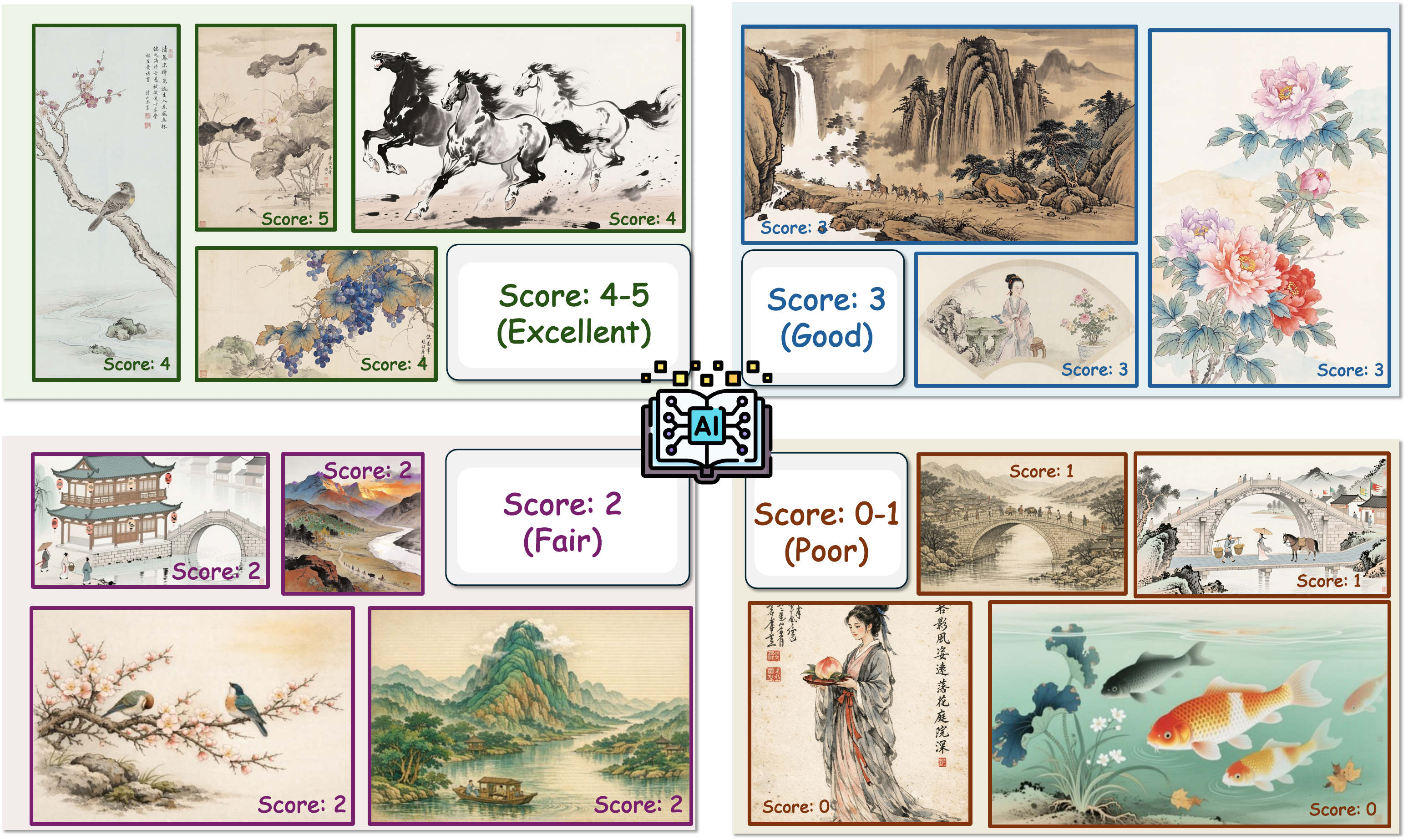}
    \vspace{-6mm}
    \caption{Generated results in different ratings groups (zoom in for details).}
    \label{fig:visual_grouped_results}
    \vspace{-3mm}
\end{figure*}

\begin{figure*}[t]
    \centering
    \includegraphics[width=1.0\linewidth]{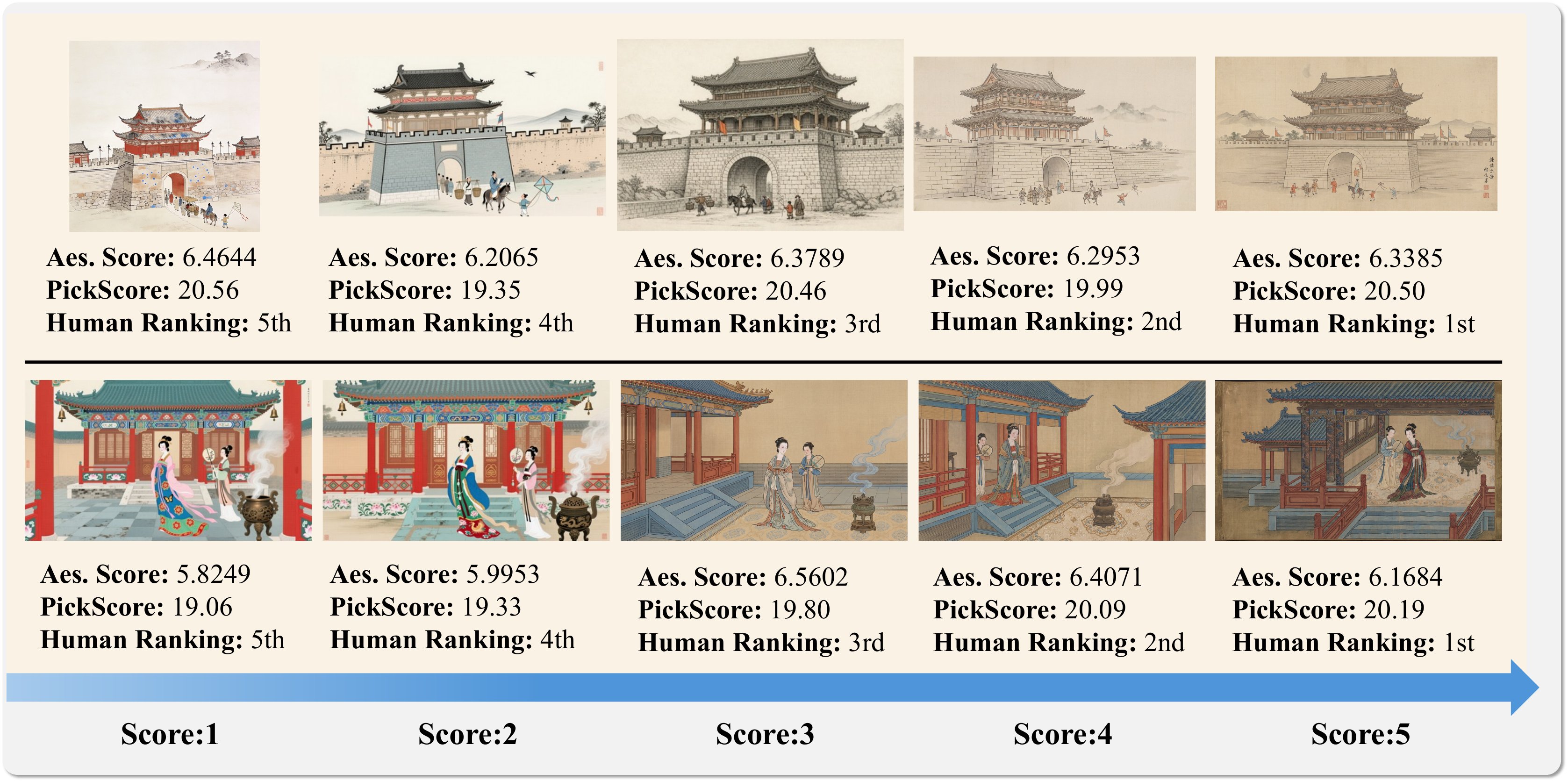}
    \vspace{-6mm}
    \caption{Generated results with the same prompt across different ratings.}
    \label{fig:visual_results}
    \vspace{-4mm}
\end{figure*}

\noindent\textbf{Visual Artistic Generation.} 
We present the visual generation results in Fig.~\ref{fig:visual_grouped_results} and Fig.~\ref{fig:visual_results}, respectively. It can be seen that images generated under the same prompt ranging from 0 to 5 points exhibit a significant gap in the quality of painting. 
Besides, PickScore~\cite{kirstain2023pick} and Aesthetic scores cannot accurately reflect human ranking of Chinese paintings. For example, the painting with the highest aesthetic score (6.4644) and PickScore (20.56) is not the first ranking from humans.
Additionally, comparing the images across different groups reveals that artworks with lower scores exhibit distinct AI artifacts or areas that clearly deviate from Chinese painting aesthetics.
More results are available in supplementary.

\begin{table*}[t]

\caption{The comparison in mIoU and Theme identification of our VLM with other models.}
\vspace{-4mm}
    \setlength{\tabcolsep}{6pt}
    \centering
    \resizebox{\textwidth}{!}{
    \setlength{\tabcolsep}{8pt}
    \begin{tabular}{lcccccc}
    \toprule
    \textbf{Model} & \textbf{Size} & \textbf{$\text{mIoU}\uparrow$}  & \textbf{$\text{BERTScore}({\text{RoI}})\uparrow$} & \textbf{$\text{Theme}\uparrow$} &  \\ \midrule
    
    Qwen3-VL-Instruct & ~4B & 0.0821 & 0.2852 & 0.7346 \\
    \textbf{HanMoVLM (SFT)} & ~4B & 0.2708 & 0.4963 & 0.8305 \\
    \textbf{HanMoVLM (RFT)} & ~4B & 0.2917 & 0.4911 & 0.8411 \\ \midrule

    Qwen3-VL-Instruct & ~8B & 0.1401 & 0.3214 & 0.7218 \\
    \textbf{HanMoVLM (SFT)} & ~8B & 0.2835 & 0.4872 & 0.8355 \\
    \textbf{HanMoVLM (RFT)} & ~8B & 0.3193 & 0.5126 & 0.8515 \\

    \bottomrule
    \end{tabular}}
    \vspace{-4mm}
    \label{tab:iou_bert_theme}
\end{table*}

\subsection{Ablation Study}
For the ablation study, we conduct experiments to investigate how the reward function improves the HanMoVLM-4B in different metrics. 
As shown in Tab.~\ref{table:ablation_study}, the model achieves the best performance on all metrics when all three rewards are combined in RFT. The results consistently show that the absence of a specific reward leads to a performance drop in its corresponding metric. For example, omitting the $R_\text{acc}$ reduces the accuracy from 42.5\% to 39.82\%. Similarly, omitting $R_\text{mIoU}$ causes the mIoU to decrease from 0.2917 to 0.2728.

%

\begin{table*}[t]

\caption{The ablation study of using different rewards.}
\vspace{-4mm}
    \setlength{\tabcolsep}{6pt}
    \centering
    \resizebox{\textwidth}{!}{
    \setlength{\tabcolsep}{12pt}
\label{table:ablation_study}
\begin{tabular}{wc{1cm} wc{1cm} wc{1cm}|wc{3cm}wc{3cm}wc{3cm}@{}} 
\toprule
\textbf{$R_\text{acc}$} & \textbf{$R_\text{BERT}$} & \textbf{$R_\text{mIoU}$} & \textbf{Accuracy (\%)$\uparrow$}& \textbf{BERTScore$\uparrow$} & \textbf{mIoU$\uparrow$} \\ \midrule
 \cmark & & \cmark  & 42.18 & 0.5529 & 0.2914    \\
 \cmark & \cmark  &  &    42.36 & 0.5492 & 0.2728       \\
              &      \cmark                & \cmark              &     39.82 & 0.5562 & 0.2902 \\ \midrule
 \cmark              & \cmark              & \cmark              &       \textbf{42.50} & \textbf{0.5568} & \textbf{0.2917}   \\ \bottomrule
\end{tabular}}
\vspace{-6mm}
\end{table*}

\subsection{Human Study}
\begin{figure*}[t]
\begin{minipage}{0.41\linewidth}
    \centering
    \includegraphics[width=1.0\linewidth]{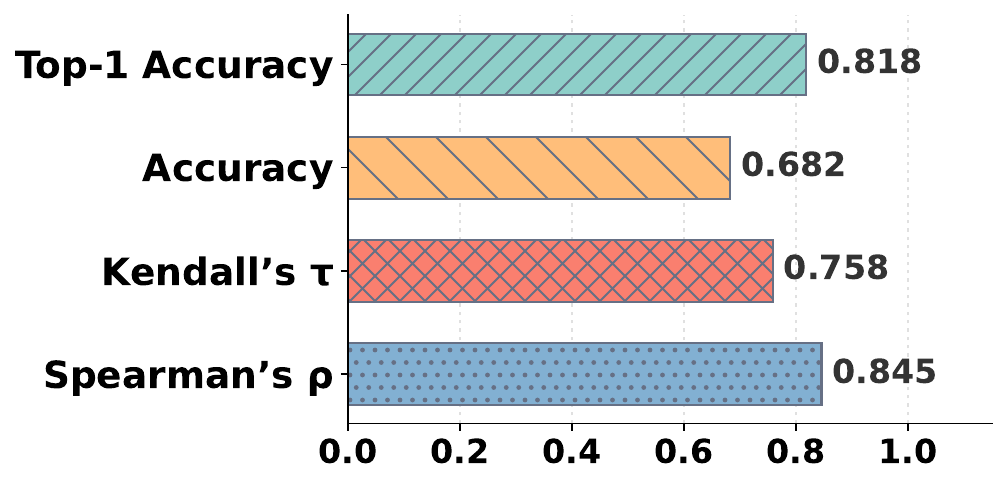}
    \caption{Consistency between HanMoVLM ratings and human rankings.}
    \label{fig:human_study}
\end{minipage}
\hspace{0.02\linewidth}
\begin{minipage}{0.58\linewidth}
    \centering
    \includegraphics[width=1.0\linewidth]{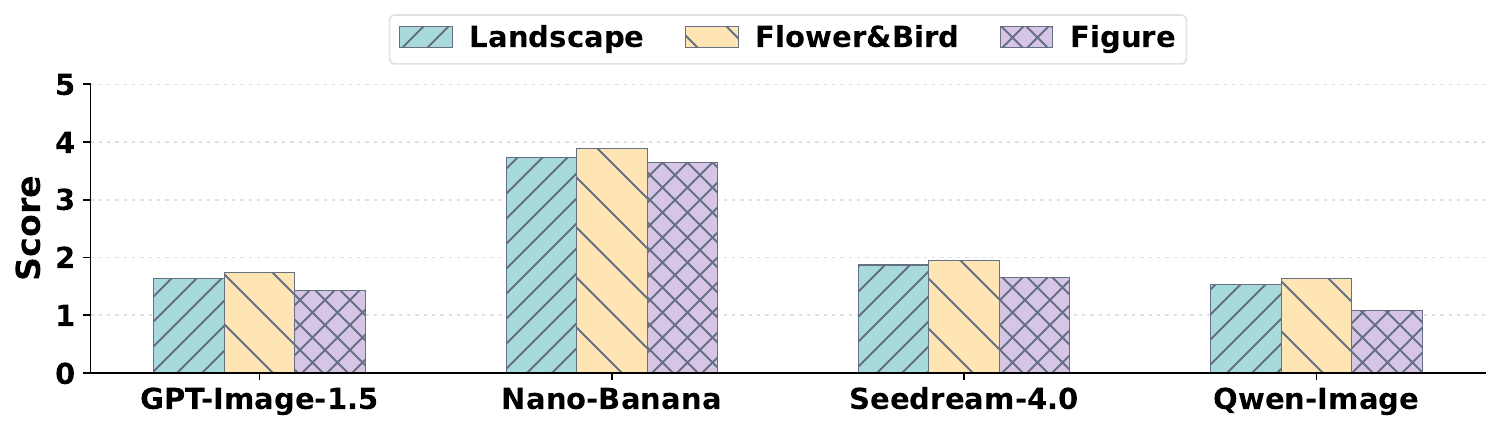}
    \caption{Comparison of T2I models across different themes.}
    \label{fig:t2i_theme}
\end{minipage}
\vspace{-6mm}
\end{figure*}

\vspace{-1mm}
To validate the consistency between our HanMoVLM's rating and human preferences, we conduct a human study where 43 researchers with backgrounds in Chinese painting are asked to rank the images sampled from the T2I model. We compare the ranking results with the scores produced by HanMoVLM.
As shown in Fig.~\ref{fig:human_study}, the ranking order obtained by HanMoVLM for the painting rating approaches that of experts. This indicates that the proposed HanMoVLM is capable of modeling expert judgments. 
Additionally, Kendall's $\tau$ and Spearman's $\rho$ correlation coefficients reached 0.758 and 0.845, respectively, suggesting that the proposed HanMoVLM aligns with human preference.



\begin{figure*}[t]
\begin{minipage}{0.46\linewidth}
    \centering
    \includegraphics[width=1.0\linewidth]{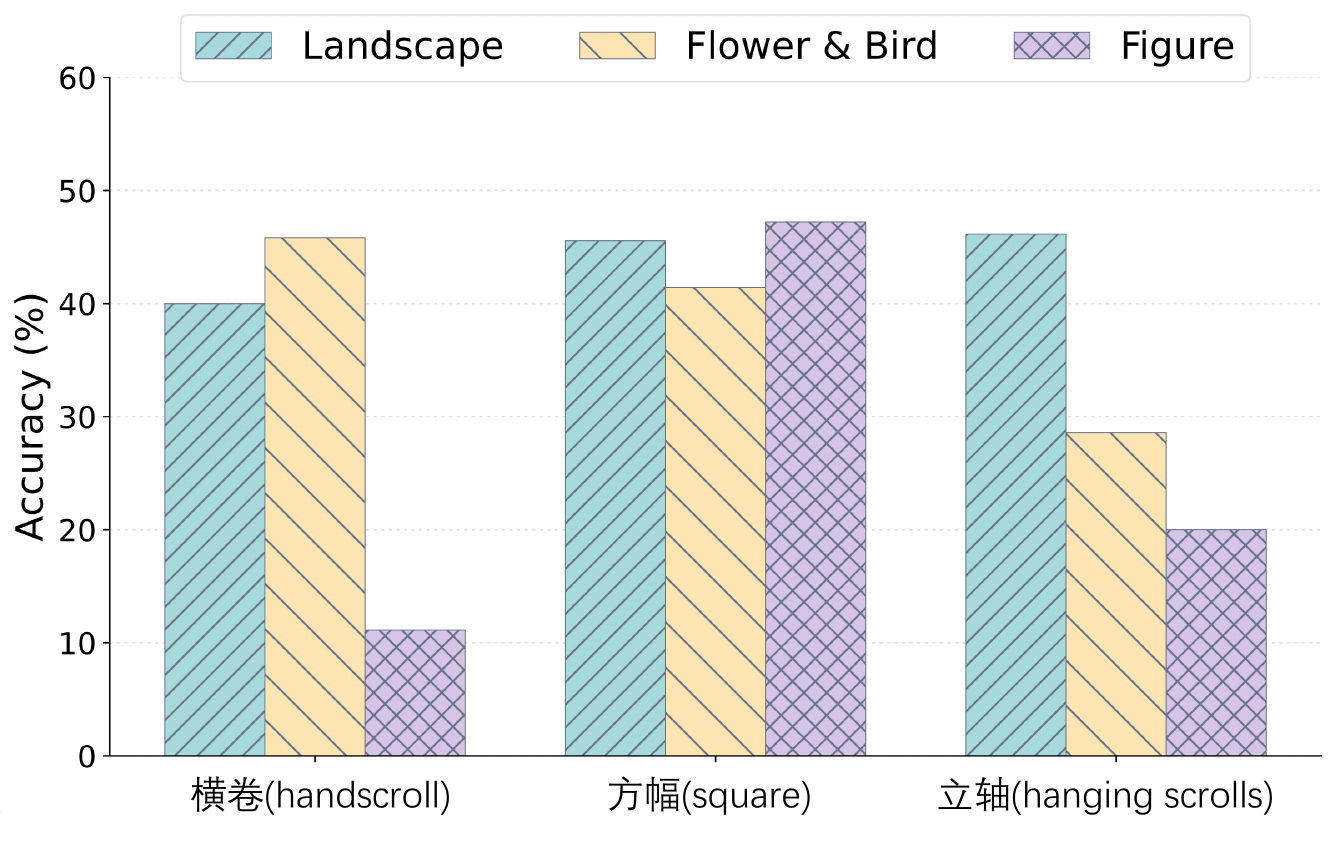}
    \caption{The effects of different types of painting.}
    \label{fig:t2i_theme_asp_score}  
\end{minipage}
\hspace{0.02\linewidth}
\begin{minipage}{0.46\linewidth}
    \centering
    \includegraphics[width=1.0\linewidth]{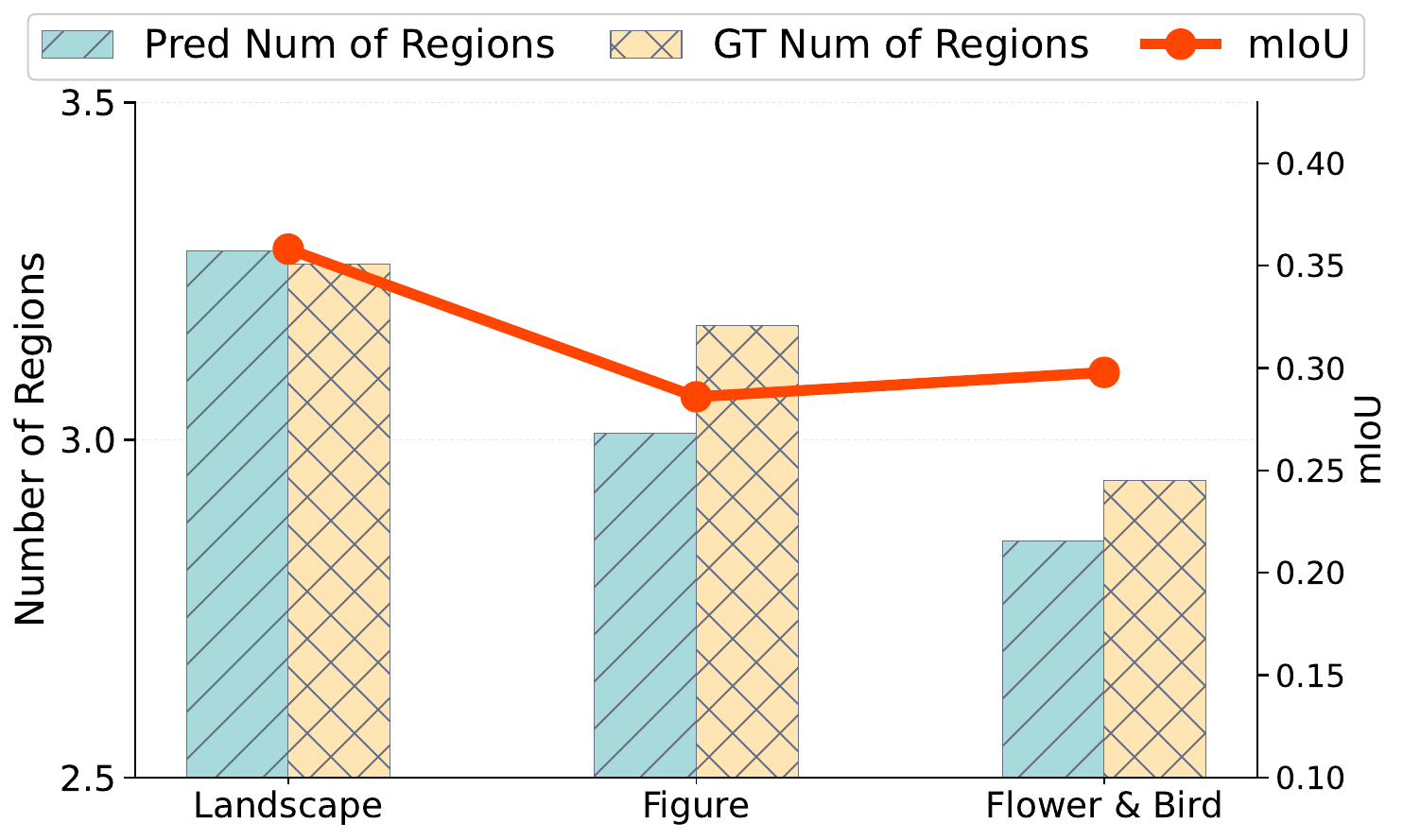}
    \caption{Number of RoI divisions and mIoU across different themes}
    \label{fig:t2i_theme_iou_score}  
\end{minipage}
\vspace{-4mm}
\end{figure*}

\subsection{Analysis}
\vspace{-1mm}
\noindent\textbf{The painting quality of different T2I models.} We conduct a study on the performance of different leading T2I models across different themes.
The results, as shown in Fig.~\ref{fig:t2i_theme}, demonstrate that Nano-Banana performs best in generating Chinese painting, significantly outperforming GPT-Image-1.5, Seedream4.0, and Qwen-Image, with a lead of approximately 2 points. Moreover, the T2I models achieve better results in Flowers\&Birds painting than in other themes.

\noindent\textbf{The effects of different type scrolls.} We present the results in different types of Chinese painting (立轴, 方幅, 横卷) in Fig.~\ref{fig:t2i_theme_asp_score}.
The 立轴~(hanging scroll) refers to the painting designed to be hung on the wall, typically with a greater height than width.
The 方幅~(square format) is a square composition that emphasizes balance and symmetry. And the 横卷~(handscroll) is a long and horizontal scroll, highlighting its extended format.
The landscape painting themes exhibit relatively stable performance across different types, while the figure painting performance shows variation. 

\noindent\textbf{The number of RoI divisions.} As shown in Fig.~\ref{fig:t2i_theme_iou_score}, the number of RoIs predicted by VLM is mostly consistent with those in the ground truth. Meanwhile, paintings focusing on specific small-scale subjects like flowers and birds require fewer RoIs compared to other types of paintings. For example, the average number of RoI divisions in the Landscape paintings is 3.25, while that in the Flowers\&Birds paintings averages around 2.8.



    



\begin{figure*}
    \centering
    \includegraphics[width=1.02\linewidth]{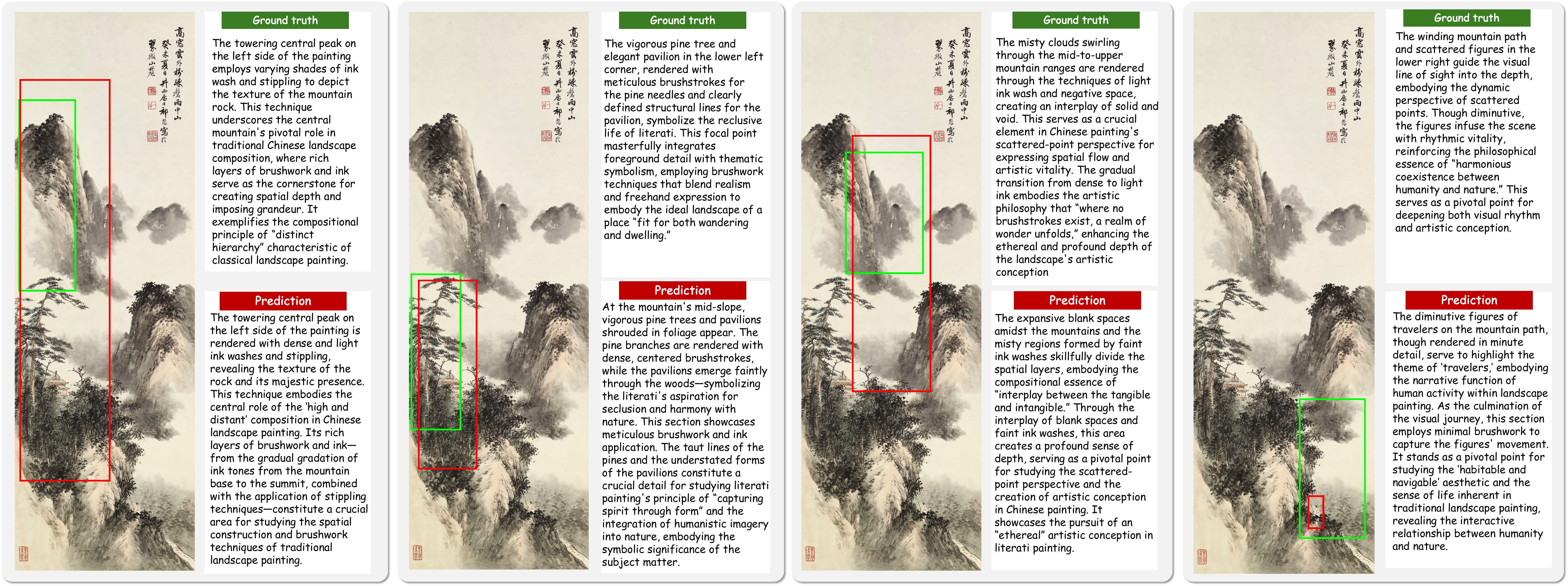}
    \vspace{-6mm}
    \caption{The case of RoI in HanMoVLM and the corresponding predicted description and gt description in HanMoVLM. Zoom in for more details.}
    \label{fig:iou_case}
    \vspace{-6mm}
\end{figure*}

\noindent\textbf{The case of RoI.}
In expert-level CoT, we guide the VLM to segment specific IoU regions. Although the mIoU value is only 0.31, as shown in the Tab.~\ref{tab:iou_bert_theme}, its ability to localize specific semantic regions remains robust. We illustrate a case with an mIoU value of only 0.31 in Fig.~\ref{fig:iou_case}. Although the ground truth boxes and predicted boxes exhibit differences, the content they describe is similar.
For instance, HanMoVLM accurately identified the objects in the painting that should be analyzed, especially since this is an extremely tall hanging scroll, and the bounding box of some targets (e.g., pedestrians) is extremely small.

\vspace{-3mm}
\section{Conclusion}
\vspace{-1mm}
In this paper, we address the problem that current generic VLMs cannot accurately evaluate the painting in the Chinese artistic domain due to artistic misalignment, the gap between understanding and evaluation, low quality data in this field.
To bridge the gap between general recognition and evaluation, we propose HanMoVLM, which leverages a structured expert-level CoT to mimic the thought of Chinese painting artists. Furthermore, we introduce a new reward mechanism to significantly enhance reasoning accuracy in the RFT.
Besides, we introduce HanMo-Bench, a novel and rigorous benchmark featuring authentic masterpieces and AI-generated works. 
Experimental results demonstrate that HanMoVLM is mostly consistent with human experts. Beyond evaluation, we showcase the model's utility as an external verifier for Test-time Scaling, effectively improving the artistic quality of generated Chinese paintings. 

%
%

\bibliographystyle{splncs04}
\bibliography{main}

\clearpage
\setcounter{page}{1}

\title{
HanMoVLM: Large Vision-Language Models for Professional Artistic Painting Evaluation
}

\author{}
\institute{}

\maketitle

\section{Implementation details}
The weights of the reward in RFT are set to $w_\text{acc}=10.0, w_\text{BERT}=2.0, w_\text{mIoU}=2.0, w_\text{format}=1.0$. For HanMoVLM training, we apply the cosine decay learning rate scheduler in both SFT and RFT. For the thinking mode, we select models built in thinking capabilities, or explicitly employ a CoT prompting with ``\textit{Let’s think step by step}''.

\begin{table*}
\caption{Comparison of ranking accuracy in different models.}
\vspace{-4mm}
    \setlength{\tabcolsep}{8pt}
    \centering
    \resizebox{\textwidth}{!}{
    \setlength{\tabcolsep}{8pt}
    \begin{tabular}{lc wc{2cm} wc{2cm} wc{2cm} wc{2cm}} 
    \toprule
        \textbf{Rating Type} & \textbf{Top-1 Accuracy$\uparrow$} & \textbf{Accuracy$\uparrow$} & \textbf{Kendall's $\tau$~$\uparrow$}& \textbf{Spearman's $\rho$~$\uparrow$} \\
    \midrule
        Aesthetic Score & 0.045 & 0.125 & -0.258 & -0.282 \\
        PickScore~\cite{kirstain2023pick} & 0.409 & 0.341 & ~0.106 & ~0.073 \\
        HPSv2.1~\cite{wu2023human} & 0.364 & 0.193 & -0.151 & -0.173 \\
        GPT-4 & 0.227 & 0.341 & ~0.197 & ~0.218 \\
        GPT-5.1 & 0.273  & 0.364 & ~0.121 & ~0.209 \\
        \rowcolor{gray!20}HanMoVLM-8B & \textbf{0.818} & \textbf{0.682} & \textbf{~0.758} & \textbf{~0.845} \\
    \bottomrule
    \end{tabular}}
    
    \label{tab:models_rankings}
\end{table*}

\section{Results}
We compare the ratings of the proposed HanMoVLM-8B with other image rating models. 
Specifically, we evaluate each model’s ability to rank paintings either by sorting their predicted scores or by directly producing a ranking. The resulting rankings are then compared with the rankings obtained from human studies, which were treated as the ground truth. The result is presented in the Tab.~\ref{tab:models_rankings}. Both Kendall’s $\tau$ and Spearman’s $\rho$ range from $-1$ to 1, where values closer to 1 indicate a higher similarity between the predicted ranking and the human-provided ranking.
The aesthetic score and PickScore have a low $\tau$ and $\rho$ value ($\le0.2$), indicating that these two rating models fail to represent the aesthetic standards in the Chinese painting field.

\clearpage

\begin{figure*}
    \centering
    \includegraphics[width=1.0\linewidth]{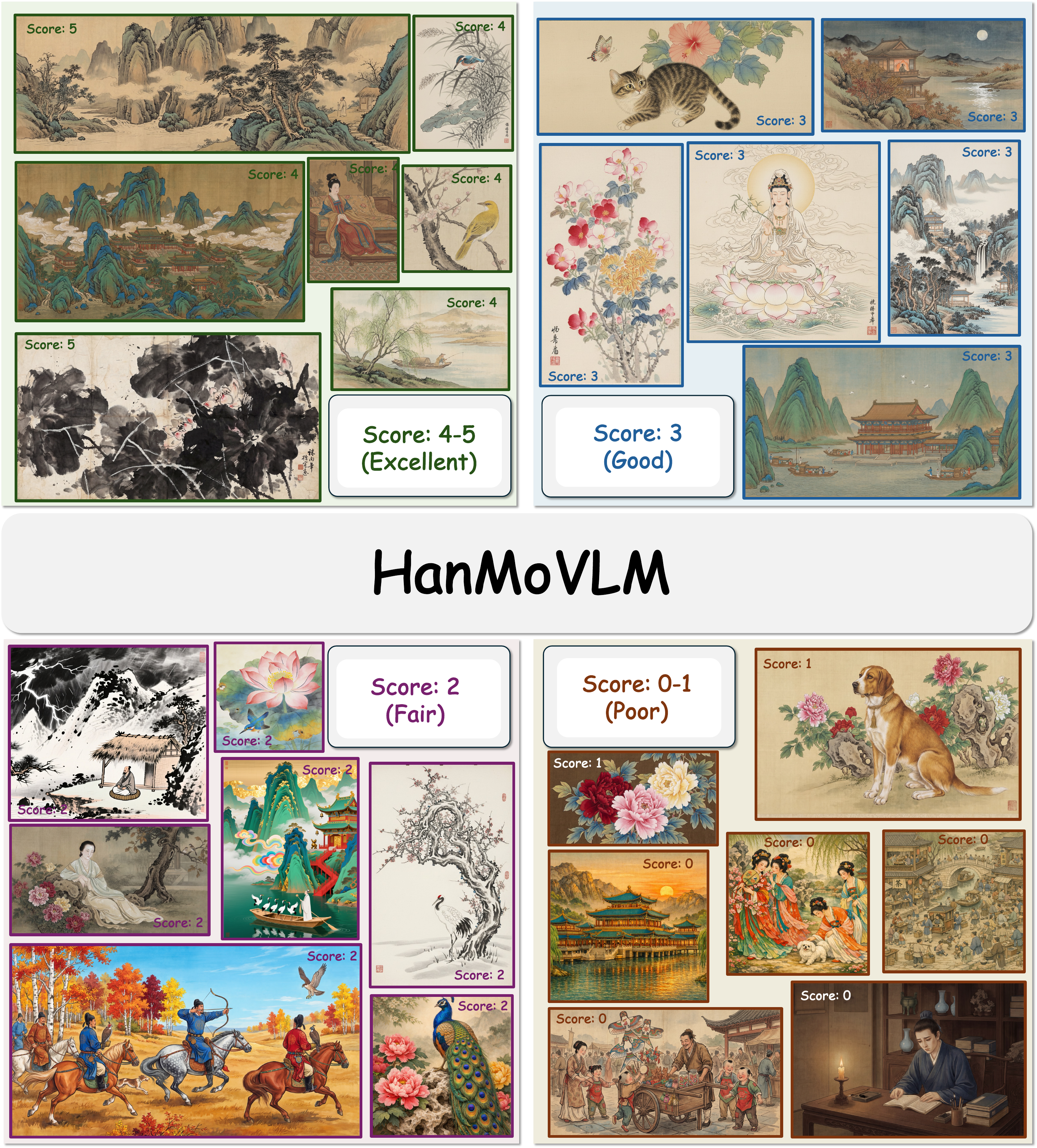}
    \vspace{-7mm}
    \caption{Visual results grouped by the score from our HanMoVLM. Zoom in for details.}
    \label{fig:viusal_result_supp}
    \vspace{-10mm}
\end{figure*}

\section{Visual Results}
\vspace{-2mm}
Fig.~\ref{fig:viusal_result_supp} shows more visual results grouped by different scores in artistic generation.
Although generative models can produce images with high-quality and remarkable details, they still fail to produce authentic Chinese paintings. Generated results with rating 0 exhibit noticeable AI artifacts, as the aesthetics of Chinese painting remain absent in generative models. Besides, the generative model incorrectly blends elements of modern painting and Western realism with the style of Chinese painting (e.g., the dog in the lower-right corner).

\clearpage
\section{Prompts}
The prompts we use in this study are shown as follows. We employ a Chinese prompt in VLM to prevent semantic gaps arising from cross-linguistic differences. We also provide an English translation for readability.
%
\subsection{Expert-Level CoT Prompt}

\begin{tcolorbox}[
        title={Expert-Level CoT Prompting (Chinese)},
        halign=left,
        valign=center,
        nobeforeafter,
        breakable,
        fontupper=\scriptsize,
    ]
你是一位中国传统绘画鉴赏专家，熟悉笔墨技法、中国画美学、艺术史与文人画理论。请对输入的国风绘画图像进行深入、专业、客观的艺术评估。让我们一步一步思考并回复

\vspace{\baselineskip}
\textbf{1. 观察这幅画作描述了什么内容，包括画面主体、构图特点以及可能的艺术风格，使用中文输出，确保描述清晰、详细}

\vspace{\baselineskip}
\textbf{2. 请明确这幅作品的具体题材分类（如：山水画、花鸟画、人物画）}

\vspace{\baselineskip}
人物画（包括历史故事、宗教人物、文人雅士、仕女、市井风俗、农耕商旅、现实人物

山水画（包括青绿山水、水墨山水、浅绛山水）

花鸟画（包括花卉、禽鸟、翎毛、蔬果、草虫、畜兽、鳞介、鱼藻

\vspace{\baselineskip}
\textbf{3. 观察画作的局部内容，确定值得分析的区域，并放大查看}

\vspace{\baselineskip}
\textbf{4. 根据画作的题材，以及该题材的评判标准，给出对这幅画的专业评价}

\vspace{\baselineskip}
人物画题材：
首先检验人物造型是否符合结构与比例，动态是否自然协调；其次重点判断人物是否具备清晰的精神气质、情绪表达与性格特征，尤其关注眼神、姿态与整体神态是否生动；同时分析线条是否具有书法性的节奏、骨力与提按变化，用笔是否支撑人物结构；再综合评估画面整体是否呈现连贯流动的生命感与精神统一性，即“气韵生动”；最终以“形神兼备、以神为主、气韵为最高标准”作为优劣判定依据，而非单纯追求形似或细节精度。

\vspace{\baselineskip}
山水画题材：
首先分析山石、树木、水体等物象的结构是否合理，皴法与笔法是否符合自然形态与山体结构；其次评估构图章法是否完整，散点透视是否自然，虚实、疏密、动静关系是否协调，留白是否有效参与空间营造；再次重点判断笔墨变化是否丰富，干湿浓淡是否层次分明，线条是否具有节奏与骨力；最后综合评估画面是否形成连贯统一的意境空间，是否呈现“可行、可望、可游、可居”的整体境界，并以“气韵生动、境由心造、笔墨当随时代”作为最高优劣判断标准，而非单纯写实或细节复杂。

\vspace{\baselineskip}
花鸟画题材：
首先分析花卉、禽鸟、草虫、鱼兽等物象的形态结构是否准确自然，比例与动态是否符合生物特征；其次重点判断物象是否具有鲜明的生命感与生动气息，是否体现自然生长节律；再次评估用笔是否灵动有力、富有节奏，墨色与设色是否协调雅致、层次分明；同时分析构图是否疏密有致、主次清晰、虚实得当；最后综合判断作品是否通过物象表达情感与人格象征，即“托物言志、以小见大”，并以“生意盎然、气韵生动、意象统一”作为最高优劣评判依据，而非单纯工细程度或色彩浓艳度。

\vspace{\baselineskip}
\textbf{5. 根据笔墨-气韵-意境三个层次进行评估。}
评判时需兼顾艺术真实感、生命感与文化深度，避免仅以画面精致度、工整度或装饰性效果作为高分依据。对于自然笔触的不完美所产生的生命感、节奏感与艺术真实，应给予正向评价；对于过度工整、机械化、装饰化的画面效果，应保持审慎态度。

原则：

笔墨：关注线条质量、运笔控制、墨色层次与结构塑造，判断其是否自然、稳定、生动，而非仅仅是否干净、精细

气韵：关注画面是否具有内在生命流动感、气场贯通性、节奏变化与动势走向，判断作品是否“活”，而非只是构图完整

意境：关注作品是否营造出诗意空间、情绪表达与文化审美高度，是否引发联想与回味，而非堆砌符号或套用意象模板

在综合评价中，意境权重高于气韵，气韵权重高于笔墨。若意境明显不足，即使笔墨精致，整体评分也不应过高。

\vspace{\baselineskip}
\textbf{6. 根据以上所有的评价，给出这张画作的综合艺术评估分数（0-5分），分数必须为整数。}

\vspace{\baselineskip}
5分：局部与整体在笔墨结构、气韵流动与精神指向上高度同构，意境深远，气韵化生，笔墨随心而不逾法，形成高度统一的生命结构与文化境界，具有不可替代的艺术原创性。

4分：关键局部笔墨精到、气脉贯通，整体节奏自然，意境清远深长，文化气息浓厚，艺术语言成熟稳定，具有鲜明而稳定的艺术品格。

3分：局部具备基本结构与气韵支撑，画面开始“有呼吸”，意境初步成立，但思想深度、文化厚度与局部—整体协同仍有限。

2分：局部描绘精细但重技巧轻生发，整体结构严谨却气韵不足，精神指向薄弱，艺术表达主要停留在技法与形式层面。

1分：构图完整，形象清晰，具备基本绘画表达，局部结构松散，笔墨僵滞，节奏单一，整体气息闭塞，尚未形成完整艺术语言。

0分：画面虽然好看、精致，但缺乏笔墨逻辑、气韵流动与意境生成，整体停留在表层视觉美感，更接近插画或装饰图。

\vspace{\baselineskip}
\textbf{按以下格式输出分数}

\vspace{\baselineskip}
\textbf{最终分数: [整数分数]}

\end{tcolorbox}


\begin{tcolorbox}[
        title={Expert-Level CoT Prompting (English)},
        halign=left,
        valign=center,
        nobeforeafter,
        fontupper=\scriptsize,
        breakable,
    ]
You are an expert in traditional Chinese painting appreciation, well-versed in brushwork techniques, Chinese painting aesthetics, art history, and literati painting theory. Please provide an in-depth, professional, and objective artistic evaluation of the submitted traditional Chinese-style painting image. Let us consider and respond step by step:

\vspace{\baselineskip}
\textbf{1. Observe and describe the content of this painting, including the main subject, compositional characteristics, and potential artistic style. Output in Chinese, ensuring the description is clear and detailed.}

\vspace{\baselineskip}
\textbf{2. Please specify the category of the theme (e.g., Landscape, Flowers\&Birds, Figure):}

\vspace{\baselineskip}
Figure (including court lady, genre painting of urban life, literati and scholars, religious figure, farming and commerce, contemporary figures, historical story)

Landscape (including Ink\&Wash, light ocher, blue\&green)

Flowers\&Birds: (Including bird\&fowl, flower, domestic animal, insect\&grass, vegetables\&fruits, scaled\&shelled creatures, fish\&aquatic plants, feather).

\vspace{\baselineskip}
\textbf{3. Observe specific details of the painting, identify areas worthy of analysis, and zoom in for inspection.}

\vspace{\baselineskip}
\textbf{4. Based on the genre and its specific criteria, provide a professional critique:}

\vspace{\baselineskip}
Figure Painting: 
First, examine if the modeling follows proper structure and proportion and if the movement is natural. Focus on whether the figure possesses a clear spiritual temperament, emotional expression, and character, paying close attention to the eyes, posture, and overall vitality. Analyze whether the linework possesses calligraphic rhythm, "bone-strength", and variations in pressure. Evaluate the "Spirit resonance" using "the integration of form and spirit, prioritizing spirit" as the standard rather than mere likeness or detail.

\vspace{\baselineskip}
Landscape Painting: 
Analyze the structure of rocks, trees, and water. Assess whether the texture strokes and brushwork align with natural forms and geological structures. Evaluate the composition, the use of "shifting perspective," and the balance between void and solid, density and sparseness. Determine if the brush and ink changes are rich (dry/wet/dark/light) and if the white space effectively creates space. The highest standard is the creation of the "Artistic Conception", a space that is "walkable, viewable, travelable, and livable"—prioritizing spirit and the idea that "brush and ink should follow the times."

\vspace{\baselineskip}
Flowers\&Birds Painting: 
Analyze the morphological structure of the subjects. Focus on whether they possess a vivid sense of life and natural growth rhythms. Evaluate if the brushwork is agile and rhythmic, and if the coloring is elegant and layered. Assess the composition for clarity of primary and secondary subjects. The ultimate judgment rests on "Expressing Ambition through Objects" and "Vividness of Life", rather than mere technical fineness or vivid colors.

\vspace{\baselineskip}
\textbf{5. Brushwork\&Ink, Spirit resonance, and Artistic conception:}

\vspace{\baselineskip}
Principle:

Brush and Ink: Focus on line quality, brush control, ink layering, and structural modeling; judge whether it is natural, stable, and vivid, rather than merely whether it is clean or refined.

\medskip

Spirit Resonance: Focus on whether the image possesses an internal sense of life-flow, a coherent "field", rhythmic variations, and momentum; judge whether the work is "alive" rather than just compositionally complete.

Artistic Conception: Focus on whether the work creates a poetic space, emotional expression, and cultural-aesthetic depth; judge whether it evokes association and reflection rather than merely piling up symbols or applying conventional image templates.

\vspace{\baselineskip}
In the comprehensive evaluation, Artistic Conception carries more weight than Spirit Resonance, and Spirit Resonance carries more weight than Brush and Ink. If the Artistic Conception is clearly insufficient, the overall score should not be high even if the brush and ink are refined.

\vspace{\baselineskip}
\textbf{6. Based on all the above evaluations, provide a comprehensive artistic assessment score (0-5), which must be an integer.}

\vspace{\baselineskip}
5 points: Local details and the whole are highly isomorphic in brushwork structure, flow of spirit, and spiritual orientation. The Artistic Conception is profound, Spirit Resonance is transformed into life, and brush and ink follow the heart without violating principles, forming a highly unified life structure and cultural realm with irreplaceable artistic originality.

4 points: Key local details are refined in brush and ink, the spiritual pulse is continuous, the overall rhythm is natural, the Artistic Conception is clear and far-reaching, the cultural atmosphere is thick, and the artistic language is mature and stable, possessing a distinct and stable artistic character.

3 points: Local parts possess basic structure and spiritual support; the image begins to "breathe," and the Artistic Conception is initially established, but depth of thought, cultural thickness, and the synergy between local and whole are still limited.

2 points: Local descriptions are fine but emphasize technique over vitality; the overall structure is rigorous but lacks Spirit Resonance, the spiritual orientation is weak, and artistic expression mainly remains at the level of technique and form.

1 point: The composition is complete and the images are clear, possessing basic pictorial expression; however, local structures are loose, brush and ink are stagnant, the rhythm is singular, and the overall atmosphere is blocked, failing to form a complete artistic language.

0 point: Although the image is attractive or exquisite, it lacks brushwork logic, the flow of spirit, and the generation of the Artistic Conception; the whole remains at the level of surface visual beauty, closer to an illustration or decorative graphic.

\vspace{\baselineskip}
\textbf{Output the score in the following format:}

\vspace{\baselineskip}
\textbf{Final rating: [Integer Score]}

\end{tcolorbox}

\vspace{-1mm}
\subsection{Prompt for Generation}
\vspace{-1mm}

\begin{tcolorbox}[
        title={Prompt for Text-to-Image models (Chinese)},
        halign=left,
        valign=center,
        nobeforeafter,
        fontupper=\scriptsize,
        breakable,
    ]
生成一个聚焦于中国画作的prompt，按格式输出20个详细的、可直接用于文本生成图像模型的prompt

1. 需要充分发挥想象，并且对出现的元素进行详细描述，但不要写出画作的名字

2. 可以指定题材，包括但不限于山水/花鸟/人物

3. 可以指定绘画手法以及色彩

4. 需要符合中国画中的笔墨-气韵-意境的特点

5. 每条prompt应该不少于150词，以中文输出

\vspace{\baselineskip}
[Prompt1]: <prompt1>

[Prompt2]: <prompt2>

[Prompt3]: <prompt3>

...

(up to 20 prompts)
\end{tcolorbox}

\begin{tcolorbox}[
        title={Prompt for Text-to-Image models (English)},
        halign=left,
        valign=center,
        nobeforeafter,
        fontupper=\scriptsize,
        breakable,
    ]
Generate 20 detailed prompts focused on Chinese painting that can be directly used for text-to-image models. Please output them according to the following format:

\vspace{\baselineskip}
1. You need to fully utilize your imagination and provide detailed descriptions of the elements involved, but do not include the titles of specific famous paintings.

2. You may specify the subject matter, including but not limited to landscapes, birds and flowers, or figures.

3. You may specify painting techniques (brushwork) and color schemes.

4. The prompts must align with the characteristics of "Brush and Ink", "Spirit Resonance", and "Artistic Conception" inherent in traditional Chinese painting.

5. Each prompt should be no less than 150 words and must be output in Chinese.

\vspace{\baselineskip}
[Prompt1]: <prompt1>

[Prompt2]: <prompt2>

[Prompt3]: <prompt3>

...

(up to 20 prompts)
\end{tcolorbox}

\subsection{CoT Construction}

\begin{tcolorbox}[
        title={Multi-turn Dialogue for CoT Construction  (Round-1)},
        halign=left,
        valign=center,
        nobeforeafter,
        fontupper=\scriptsize,
        breakable,
    ]
请尽可能详细描述这幅图像的内容，包括画面主体、构图特点以及可能的艺术风格，使用中文输出，确保描述清晰、详细，不少于100个词。

\vspace{\baselineskip}
基于前面的描述，请明确这幅作品的具体题材分类（如：山水画、花鸟画、人物画等）。

\vspace{\baselineskip}
人物画（包括历史故事、宗教人物、文人雅士、仕女、市井风俗、农耕商旅、现实人物)

山水画（包括青绿山水、水墨山水、浅绛山水）

花鸟画（包括花卉、禽鸟、翎毛、蔬果、草虫、畜兽、鳞介、鱼藻

\vspace{\baselineskip}
(English version)

Please describe the content of this image in as much detail as possible, including the main subject, compositional features, and possible artistic style. Provide the description in Chinese, ensuring it is clear and detailed, with a minimum of 100 words.

Based on the description, specify the category of the theme (e.g., Landscape, Flowers\&Birds, Figure):

\vspace{\baselineskip}
Figure (including court lady, genre painting of urban life, literati and scholars, religious figure, farming and commerce, contemporary figures, historical story)

Landscape (including Ink\&Wash, light ocher, blue\&green)

Flowers\&Birds: (Including bird\&fowl, flower, domestic animal, insect\&grass, vegetables\&fruits, scaled\&shelled creatures, fish\&aquatic plants, feather).

\end{tcolorbox}

\begin{tcolorbox}[
        title={Multi-turn Dialogue for CoT Construction  (Round-2)},
        halign=left,
        valign=center,
        nobeforeafter,
        fontupper=\scriptsize,
        breakable,
    ]
基于上述描述，请你从艺术与视觉结构角度分析画面内容，识别出最具研究价值的感兴趣区域（Region of Interest, ROI），并为每个区域提供精确的 bounding box。

你需要仔细看一下这张图像的内容，然后根据图像的内容分析出具体需要有几个感兴趣区域。

\vspace{\baselineskip}
分析要求：
1. 综合考虑中国画的构图方式（如：主次关系、散点透视）、笔墨技法、题材象征意义。

2. 感兴趣区域可以包括：主要描绘对象、视觉中心、具有显著笔墨特征的局部，视觉中心或视觉动线的关键节点。

3. 每个感兴趣区域需给出明确的语义说明。

4. 不要分析题跋章印或者文字部分，聚焦在画作视觉元素本身

\vspace{\baselineskip}
输出格式要求：

请以 JSON 格式输出结果，使用归一化的像素坐标（范围0-1），注意精度尽可能准确到小数点后5位，坐标原点为图像左上角。

确保输出为合法 JSON，输出一个 JSON 对象。

使用中文。

\begin{verbatim}
{
  "height": {height},
  "width": {width},
  "num_regions": N,
  "regions_of_interest": [
    {
      "label": "区域名称",
      "description": "该区域在中国画中的艺术或研究意义",
      "bounding_box": {
        "x_min": x1,
        "y_min": y1,
        "x_max": x2,
        "y_max": y2
      }
    }
  ]
}
\end{verbatim}

\vspace{\baselineskip}
(English version)

Based on the description above, please analyze the content of the image from an artistic and visual structure perspective, identify the Regions of Interest (ROI) with the greatest research value, and provide precise bounding boxes for each region.

Please examine the content of this image carefully, and then analyze how many specific Regions of Interest (ROI) are required based on the visual information.

\vspace{\baselineskip}
Requirements:

1. Comprehensively consider the compositional methods of Chinese painting (e.g., primary-secondary relationships, shifting perspective), brush and ink techniques, and the symbolic significance of the subject matter.

2. Regions of Interest may include: the primary subjects, the visual center, localized areas with significant brushwork characteristics, and key nodes of the visual center or visual flow.

3. Provide a clear semantic description for each Region of Interest.

4. Do not analyze inscriptions, colophons, or seals; focus exclusively on the visual elements of the artwork itself.

\vspace{\baselineskip}

Output Format Requirements:

Please output the results in JSON format using normalized pixel coordinates (range 0-1). Ensure the precision is as accurate as possible to 5 decimal places, with the coordinate origin at the top-left corner of the image.

Ensure the output is a valid JSON object.

Use Chinese for the descriptions.

\begin{verbatim}
{
  "height": {height},
  "width": {width},
  "num_regions": N,
  "regions_of_interest": [
    {
      "label": "Label of the regions",
      "description": "The artistic significance of this region in Chinese painting.",
      "bounding_box": {
        "x_min": x1,
        "y_min": y1,
        "x_max": x2,
        "y_max": y2
      }
    }
  ]
}
\end{verbatim}

\end{tcolorbox}

\begin{tcolorbox}[
        title={Multi-turn Dialogue for CoT Construction  (Round-3)},
        halign=left,
        valign=center,
        nobeforeafter,
        fontupper=\scriptsize,
        breakable,
    ]
基于前面的描述，根据该题材所属的特定审美标准，对作品进行简要评价。

\vspace{\baselineskip}
人物画题材：

首先检验人物造型是否符合结构与比例，动态是否自然协调；其次重点判断人物是否具备清晰的精神气质、情绪表达与性格特征，尤其关注眼神、姿态与整体神态是否生动；同时分析线条是否具有书法性的节奏、骨力与提按变化，用笔是否支撑人物结构；再综合评估画面整体是否呈现连贯流动的生命感与精神统一性，即“气韵生动”；最终以“形神兼备、以神为主、气韵为最高标准”作为优劣判定依据，而非单纯追求形似或细节精度。

\vspace{\baselineskip}
山水画题材：

首先分析山石、树木、水体等物象的结构是否合理，皴法与笔法是否符合自然形态与山体结构；其次评估构图章法是否完整，散点透视是否自然，虚实、疏密、动静关系是否协调，留白是否有效参与空间营造；再次重点判断笔墨变化是否丰富，干湿浓淡是否层次分明，线条是否具有节奏与骨力；最后综合评估画面是否形成连贯统一的意境空间，是否呈现“可行、可望、可游、可居”的整体境界，并以“气韵生动、境由心造、笔墨当随时代”作为最高优劣判断标准，而非单纯写实程度或细节复杂度。

\vspace{\baselineskip}
花鸟画题材：

首先分析花卉、禽鸟、草虫、鱼兽等物象的形态结构是否准确自然，比例与动态是否符合生物特征；其次重点判断物象是否具有鲜明的生命感与生动气息，是否体现自然生长节律；再次评估用笔是否灵动有力、富有节奏，墨色与设色是否协调雅致、层次分明；同时分析构图是否疏密有致、主次清晰、虚实得当；最后综合判断作品是否通过物象表达情感与人格象征，即“托物言志、以小见大”，并以“生意盎然、气韵生动、意象统一”作为最高优劣评判依据，而非单纯工细程度或色彩浓艳度。

\vspace{\baselineskip}
(English version)

Based on the preceding description, provide a brief evaluation of the work according to the specific aesthetic standards applicable to its genre.

\vspace{\baselineskip}
Figure Painting: 

First, examine if the modeling follows proper structure and proportion and if the movement is natural. Focus on whether the figure possesses a clear spiritual temperament, emotional expression, and character, paying close attention to the eyes, posture, and overall vitality. Analyze whether the linework possesses calligraphic rhythm, "bone-strength", and variations in pressure. Evaluate the "Spirit resonance" using "the integration of form and spirit, prioritizing spirit" as the standard rather than mere likeness or detail.

\vspace{\baselineskip}
Landscape Painting: 

Analyze the structure of rocks, trees, and water. Assess whether the texture strokes and brushwork align with natural forms and geological structures. Evaluate the composition, the use of "shifting perspective," and the balance between void and solid, density and sparseness. Determine if the brush and ink changes are rich (dry/wet/dark/light) and if the white space effectively creates space. The highest standard is the creation of the "Artistic Conception", a space that is "walkable, viewable, travelable, and livable"—prioritizing spirit and the idea that "brush and ink should follow the times."

\vspace{\baselineskip}
Flowers\&Birds Painting: 

Analyze the morphological structure of the subjects. Focus on whether they possess a vivid sense of life and natural growth rhythms. Evaluate if the brushwork is agile and rhythmic, and if the coloring is elegant and layered. Assess the composition for clarity of primary and secondary subjects. The ultimate judgment rests on "Expressing Ambition through Objects" and "Vividness of Life", rather than mere technical fineness or vivid colors.

\end{tcolorbox}

\begin{tcolorbox}[
        title={Multi-turn Dialogue for CoT Construction  (Round-4)},
        halign=left,
        valign=center,
        nobeforeafter,
        fontupper=\scriptsize,
        breakable,
    ]

请从 笔墨、气韵、意境 三个层次进行逐级分析。三者之间存在递进关系：

评判时需兼顾艺术真实感、生命感与文化深度，避免仅以画面精致度、工整度或装饰性效果作为高分依据。对于自然笔触的不完美所产生的生命感、节奏感与艺术真实，应给予正向评价；对于过度工整、机械化、装饰化的画面效果，应保持审慎态度。

\vspace{\baselineskip}
原则：

笔墨：关注线条质量、运笔控制、墨色层次与结构塑造，判断其是否自然、稳定、生动，而非仅仅是否干净、精细

气韵：关注画面是否具有内在生命流动感、气场贯通性、节奏变化与动势走向，判断作品是否“活”，而非只是构图完整

意境：关注作品是否营造出诗意空间、情绪表达与文化审美高度，是否引发联想与回味，而非堆砌符号或套用意象模板

\vspace{\baselineskip}
在综合评价中，意境权重高于气韵，气韵权重高于笔墨。若意境明显不足，即使笔墨精致，整体评分也不应过高。

\vspace{\baselineskip}
请按以下顺序进行分析：

1. 笔墨分析：用一段自然语言，评估线条、运笔、墨色变化与造型结构，指出优点与不足。

2. 气韵分析：评估画面整体生命感、动势、节奏与气场流动，判断其是否生动、有呼吸感。

3. 意境分析：评估作品是否营造出明确审美境界，是否具有情绪感染力与文化韵味，是否在有限视觉信息中，构建出超出画面本身的空间感、情绪感与联想空间

\vspace{\baselineskip}
按以下格式输出：

笔墨分析: [笔墨分析结果]

气韵分析: [气韵分析结果]

意境分析: [意境分析结果]

\vspace{\baselineskip}
(English version)

Please perform a step-by-step analysis across three levels: Brush and Ink, Spirit Resonance, and Artistic Conception. A progressive relationship exists between these three:

The evaluation must balance artistic authenticity, the sense of vitality, and cultural depth. Avoid using visual refinement, neatness, or decorative effects as the sole basis for high scores. Positive recognition should be given to the sense of life, rhythm, and artistic truth arising from the imperfections of natural brushstrokes; conversely, a cautious stance should be maintained toward visual effects that are overly meticulous, mechanized, or purely decorative.

\vspace{\baselineskip}
Principle:

Brush and Ink: Focus on line quality, brush control, ink layering, and structural modeling; judge whether it is natural, stable, and vivid, rather than merely whether it is clean or refined.

Spirit Resonance: Focus on whether the image possesses an internal sense of life-flow, a coherent "field", rhythmic variations, and momentum; judge whether the work is "alive" rather than just compositionally complete.

Artistic Conception: Focus on whether the work creates a poetic space, emotional expression, and cultural-aesthetic depth; judge whether it evokes association and reflection rather than merely piling up symbols or applying conventional image templates. In the comprehensive evaluation, Artistic Conception carries more weight than Spirit Resonance, and Spirit Resonance carries more weight than Brush and Ink. If the Artistic Conception is clearly insufficient, the overall score should not be high even if the brush and ink are refined.

In a comprehensive evaluation, the weight of Artistic Conception is higher than Spirit Resonance, and the weight of Spirit Resonance is higher than Brush and Ink. If the Artistic Conception is clearly insufficient, the overall score should not be high, even if the brush and ink are refined.

\vspace{\baselineskip}
Please perform the analysis in the following order:

Brush and Ink Analysis: Use natural language to evaluate line quality, brushwork, ink variations, and structural modeling, noting both strengths and deficiencies.

Spirit Resonance Analysis: Evaluate the overall sense of life, momentum, rhythm, and the flow of the "field" within the frame, judging whether it is vivid and possesses a sense of "breathing."

Artistic Conception Analysis: Evaluate whether the work creates a clear aesthetic conception, possesses emotional resonance and cultural charm, and whether it constructs a sense of space, emotion, and association that transcends the visual information of the canvas itself.

\vspace{\baselineskip}
Output in the following format:

Brush and Ink Analysis: [Results of Brush and Ink Analysis]

Spirit Resonance Analysis: [Results of Spirit Resonance Analysis]

Artistic Conception Analysis: [Results of Artistic Conception Analysis]

\end{tcolorbox}

\begin{tcolorbox}[
        title={Multi-turn Dialogue for CoT Construction  (Round-5)},
        halign=left,
        valign=center,
        nobeforeafter,
        fontupper=\scriptsize,
        breakable,
    ]

在综合图像描述、基于题材的艺术分析和评价、RoI感兴趣区域的描述和评价以及三层次分析后，给出这张画作最终的综合艺术评估分数（0-5分），分数必须为整数。

\vspace{\baselineskip}

5分：局部与整体在笔墨结构、气韵流动与精神指向上高度同构，意境深远，气韵化生，笔墨随心而不逾法，形成高度统一的生命结构与文化境界，具有不可替代的艺术原创性。

4分：关键局部笔墨精到、气脉贯通，整体节奏自然，意境清远深长，文化气息浓厚，艺术语言成熟稳定，具有鲜明而稳定的艺术品格。

3分：局部具备基本结构与气韵支撑，画面开始“有呼吸”，意境初步成立，但思想深度、文化厚度与局部—整体协同仍有限。

2分：局部描绘精细但重技巧轻生发，整体结构严谨却气韵不足，精神指向薄弱，艺术表达主要停留在技法与形式层面。

1分：构图完整，形象清晰，具备基本绘画表达，局部结构松散，笔墨僵滞，节奏单一，整体气息闭塞，尚未形成完整艺术语言。

0分：画面虽然好看、精致，缺乏笔墨逻辑、气韵流动与意境生成，整体停留在表层视觉美感，更接近插画或装饰图。

\vspace{\baselineskip}
按以下格式输出，不要输出额外内容。

最终分数: [整数分数]

\vspace{\baselineskip}
(English version)

5 points: Local details and the whole are highly isomorphic in brushwork structure, flow of spirit, and spiritual orientation. The Artistic Conception is profound, Spirit Resonance is transformed into life, and brush and ink follow the heart without violating principles, forming a highly unified life structure and cultural realm with irreplaceable artistic originality.

4 points: Key local details are refined in brush and ink, the spiritual pulse is continuous, the overall rhythm is natural, the Artistic Conception is clear and far-reaching, the cultural atmosphere is thick, and the artistic language is mature and stable, possessing a distinct and stable artistic character.

3 points: Local parts possess basic structure and spiritual support; the image begins to "breathe," and the Artistic Conception is initially established, but depth of thought, cultural thickness, and the synergy between local and whole are still limited.

2 points: Local descriptions are fine but emphasize technique over vitality; the overall structure is rigorous but lacks Spirit Resonance, the spiritual orientation is weak, and artistic expression mainly remains at the level of technique and form.

1 point: The composition is complete and the images are clear, possessing basic pictorial expression; however, local structures are loose, brush and ink are stagnant, the rhythm is singular, and the overall atmosphere is blocked, failing to form a complete artistic language.

0 point: Although the image is attractive or exquisite, it lacks brushwork logic, the flow of spirit, and the generation of the Artistic Conception; the whole remains at the level of surface visual beauty, closer to an illustration or decorative graphic.

Ouput the score in the following format, do not output irreverent content.

Final rating: [Integer Score]

\end{tcolorbox}

\end{CJK*}
\end{document}